\pdfoutput=1
\ifdefined\pdfsuppressptexinfo\pdfsuppressptexinfo=-1\fi
\ifdefined\pdfinfoomitdate\pdfinfoomitdate=1\fi
\ifdefined\pdftrailerid\pdftrailerid{}\fi
\documentclass[10pt,twocolumn]{article}
\usepackage[a4paper,margin=2cm,columnsep=0.6cm]{geometry}
\usepackage[T1]{fontenc}
\usepackage{lmodern}
\usepackage{amssymb,amsmath}
\usepackage{microtype}
\usepackage{pgfplots}\pgfplotsset{compat=1.18}
\usetikzlibrary{positioning,arrows.meta,fit,calc,backgrounds}
\usepackage{booktabs,array,tabularx,multirow}
\usepackage{enumitem}
\usepackage{xcolor}
\usepackage{caption}
\usepackage[hidelinks,pdftitle={The Copy Ceiling: An Input-Exposure Control for Ontology-Grounded Generation over Curated Corpora},pdfauthor={John O'Hare},pdfcreator={},pdfproducer={},pdfsubject={},pdfkeywords={}]{hyperref}
\usepackage[backend=bibtex,style=numeric-comp,sorting=nyt,maxnames=4,giveninits=true]{biblatex}
\definecolor{teal0}{RGB}{0,110,110}
\definecolor{burnt}{RGB}{192,80,0}
\definecolor{ink}{RGB}{25,28,32}
\definecolor{softgreen}{RGB}{46,139,87}
\definecolor{grayln}{RGB}{200,200,200}
\color{ink}
\hypersetup{colorlinks=true,linkcolor=teal0,citecolor=teal0,urlcolor=teal0}
\pgfplotsset{
  every axis/.append style={font=\small,axis line style={grayln},tick style={grayln},
    grid=major,major grid style={grayln!35},label style={color=ink},tick label style={color=ink}},
  rawstyle/.style={fill=burnt!85,draw=burnt},
  scafstyle/.style={fill=teal0!85,draw=teal0},
  copystyle/.style={fill=gray!35,draw=gray!60},
}

\newcommand{\gainpos}[1]{\textcolor{softgreen}{$+#1$}}
\newcommand{\gainneg}[1]{\textcolor{burnt}{$#1$}}

\title{\vspace{-1.2cm}\bfseries The Copy Ceiling: An Input-Exposure Control\\
for Ontology-Grounded Generation over Curated Corpora}
\author{Dr John O'Hare\\ \small DreamLab AI, The Loom / VisionFlow neurosymbolic stack}
\date{29 September 2026}

\begin{document}
\maketitle

\begin{abstract}
\noindent
We built a node that grounds a replaceable language model in a maintained ontology corpus, then asked what its successful-looking evaluation could support. Across ten models, grounding raised target-name recall from 0.265 unaided to about 0.92. A copy baseline, the recall a verbatim copy of the shown context already achieves, scores 0.964, and every model sits 0.022 to 0.067 below it. Copying therefore scores higher on this limited recall measure, which does not assess whether answers are better. The comparison tests what a recall score establishes; it does not test whether reasoning occurred, because a reasoned answer and a copy score alike when the answer name is already in context. We report exposure accounting (four counts classifying each gold item by whether the context exposed it and the answer recovered it) and a model-judged audit of 423 sampled item observations. A separate paired production study found a model-judged quality gain of $+0.27$ $[+0.11,+0.45]$ on a 0--5 scale. Operational studies found failures that recall alone would not show: rephrasing questions out of the graph's vocabulary cut exposure from 0.964 to 0.328, yet the absence-keyed fallback would have fired on only 2 of 506; and inserting extracted facts degraded judged pages in every arm, so that step was disabled. Five-arm controls show that any well-formed on-corpus block beats no context but do not establish that the specific content matters, and no matched comparison against flat-text retrieval was run. The corpus is public and largely LLM-generated, which establishes neither training exposure nor novelty. Each study has its own outcome measure. Where gold derives from the injected corpus, we recommend reporting the accounting beside quality judgements, not in place of them.
\end{abstract}

\section{Introduction}\label{sec:intro}
We built a serving node so that a body of maintained, research-directed knowledge could be answered through a language model the operator can replace. The corpus was developed to support the VisionFlow stack and the author's research; the node retrieves from it, injects a bounded block of it into the prompt, and delegates generation to whichever model sits behind an OpenAI-compatible fa\c{c}ade. Retrieval-augmented generation is the established way to supply knowledge a model does not reliably hold~\cite{lewis2020rag,gao2023ragsurvey,mallen2023trust}, and graph-structured retrieval is a common refinement of it~\cite{edge2024graphrag,guo2024lightrag,baek2023kaping,pan2024unifying}. This paper is an empirical systems evaluation of that node. Its reusable output is a diagnostic and a reporting procedure, not a new model or retrieval method.

\paragraph{How the control came to be built.} We deployed the node first, and the conventional measurement gave a large uplift over the same model answering unaided, of the kind this literature routinely reports. We then put a narrower question to our own result: what would a verbatim copy of the text we had already shown the model score on the same questions? That copy baseline sat above the grounded system on every model we swept, which changed what the uplift could be taken to support. It did not settle whether the node was useful, so we measured usefulness separately: first with a paired production study judged on answer quality, then with negative controls, stress tests and write-path studies asking questions recall cannot answer. The protocol below writes the copy check down so that another team can run it before reporting an uplift figure.

\paragraph{A fair question, and its limit.} Suppose the context says that Project~A uses Module~B and that Module~B requires Component~C, and the question asks which component Project~A needs. A good answer may have to follow both relations, yet ``Component~C'' already appears in the context, so a reasoned answer and a verbatim copy of the context score the same on recall of the answer name. \emph{The copy comparison tests what a recall score establishes; it does not test whether reasoning occurred.} A grounded score below the copy baseline therefore does not show that a model failed to reason, and a score above it would not show graph deduction, since memory, guessing and matcher asymmetry produce the same sign. What the comparison does show is how much of a recall uplift the names already in front of the model explain.

\textbf{Exposure accounting} makes that comparison explicit: each gold item is classified by whether the shown text exposed it and whether the answer recovered it, and the four resulting counts are reported. The \emph{copy ceiling}\footnote{The term \emph{copy ceiling} is used independently and in an unrelated sense by~\textcite{copyasdecode2026} (a decoding-layer editing mechanism, where it denotes token reachability under a line-level copy primitive; that work is withdrawn), noted here only to head off a terminology collision.} is the accounting's scalar, called a copy baseline wherever the explanation matters. It is the recall a no-op extractor of exactly the shown text achieves: a reference point, \emph{not} an upper bound on what an answer may score. The signed \emph{gain over copy} is grounded recall minus that baseline. The counts depend only on a surface matcher applied to fixed text, so they are deterministic and need no judge. Throughout, \emph{delivery} means only that an exposed gold name reappears in the answer, scored as recall of exposed target names and nothing more. The construction generalises the input-only baselines used to interpret reading comprehension~\cite{kaushik2018reading} and natural-language inference~\cite{gururangan2018artifacts,poliak2018hypothesis}.

\paragraph{Populations and instruments.} The studies differ in questions, models and outcome measures, and none is pooled with another (Table~\ref{tab:instruments}). A ten-model sweep of 510 graph-derived questions measures lexical recall of gold target names. A stratified audit of 423 sampled item observations from that sweep asks a model judge whether a credited name is asserted in the relation the question asked for. A paired production study of 114 questions scores answers from the served node against a reference on a 0--5 quality scale, again with a model judge. Negative controls on 57 of those questions perturb the injected block. A paraphrase set, an analysis of corpus targets the scaffold omitted, and a write-path case study on podcast extraction each measure an operational property of the deployed path.

\paragraph{What we find.} Unaided, the ten models average 0.265 recall of gold target names on in-domain questions (range $0.151$--$0.375$); grounded, about $0.92$. The copy baseline is $0.964$, so gain over copy is negative for every model ($-0.067$ to $-0.022$), and recovery of gold the context did not expose totals three items in 11{,}360 item$\times$model observations (the same 1{,}136 target instances scored under each of ten models, not independent facts). On this measure the models differ mainly in how many exposed target names their answers omit, which is not a ranking on answer quality. The production study found a model-judged quality benefit ($+0.27$ on the 0--5 scale). The operational studies found three failures that the recall figures could not show: a retrieval fallback that does not fire when rephrasing defeats lexical retrieval, lower recovery of corpus targets that the scaffold omitted, and a page-integration step that degraded every page it touched and was disabled.

\paragraph{Bounded contribution.} We contribute a cheap, judge-free diagnostic for recall uplift, a four-count reporting convention, a seed-disjoint placebo with measured exposure, a stratified audit that prices what the lexical matcher gives up, and a set of deployment findings with the decisions they changed. The diagnostic is available once questions, gold and candidate contexts exist; it does not price user benefit, curation, governance, model substitution or reasoning. Nor does the paper compare graph-structured grounding with strong flat-text retrieval: no matched comparison was run, and the controls are no substitute for one. A graph may contribute through retrieval, organisation, versioning and governance as well as through inference, and this study does not isolate the value or cost of each. The closest instruments, RAGChecker's context utilisation~\cite{ragchecker2024}, the reader--context presence diagnostic~\cite{readercontext2026} and Sufficient Context~\cite{sufficientcontext2025}, are compared in \S\ref{sec:related} and \S\ref{sec:ceiling-prior}; we claim narrower ground than each.

\paragraph{Release.} The served node, the ten-model sweep rows and the harness that generated them, the exposure decomposition, the production paired study and both of its judge passes, both negative-control cohorts with their completions, scores and (for the five-arm rerun) the injected blocks and their measured exposure, the semantic re-score, the paraphrase stress set and the semantic audit (sample, cached judge calls, summariser and the blind sample prepared for a human pass) are under version control, with per-study seeds named in the released scripts. This manuscript is tagged \texttt{paper-v10}; its table-to-artefact manifest is \texttt{MANIFEST.md} at \url{https://github.com/DreamLab-AI/loom/tree/paper-v10/docs/research/gain-over-copy-paper}, and every table and figure resolves to an artefact listed there. A reader can check every reported number against saved rows but cannot regenerate the scaffolds, because the corpus-derived scaffold index is not released (\S\ref{sec:limits}).

\section{Related Work}\label{sec:related}
\paragraph{Knowledge-graph grounding.}
The grounding lineage runs from RAG~\cite{lewis2020rag} through graph-structured retrieval~\cite{edge2024graphrag,guo2024lightrag} and zero-shot knowledge-graph triple prompting~\cite{baek2023kaping} to roadmaps and surveys~\cite{pan2024unifying,peng2024graphragsurvey}. GraphRAG addresses collections outside a model's training data~\cite{edge2024graphrag}, while enterprise variants highlight amortised curation~\cite{min2025practicalgraphrag}. Published work typically injects \emph{instance} triples or opaque encodings, including community summaries~\cite{edge2024graphrag} and GNN soft-prompts over retrieved subgraphs~\cite{he2024gretriever}; we instead inject human-readable, schema-level content: class definitions, typed relations and taxonomy. None of these graph-grounding studies reports a standing input-exposure control beside its uplift figures, so their headline uplift is not re-centred on content already present in context. Where retrieval coverage is reported, a reader can approximate the subtraction from the published quantities.

\paragraph{Baselines that reframe a result.}
Simple controls can overturn headline results: strong RAG baselines match elaborate pipelines under equal token budgets~\cite{laitenberger2025stronger}; controlled comparisons reverse earlier ``RAG wins'' findings and add a confidence router~\cite{li2024ragorlongcontext}; and apparent argument comprehension can reflect spurious cues~\cite{niven2019probing}. The confidence router of~\textcite{li2024ragorlongcontext} and selective retrieval~\cite{asai2024selfrag,jeong2024adaptiverag,mallen2023trust} provide precedents for our injection gate. Those methods choose whether to retrieve; ours chooses whether to inject.

\paragraph{Input-only baselines and context-adherence metrics.}
The copy ceiling belongs to this tradition. Passage-only, question-only~\cite{kaushik2018reading} and hypothesis-only baselines~\cite{gururangan2018artifacts,poliak2018hypothesis} reveal shortcuts present in inputs; unlike these discrete task classifiers, the copy ceiling is continuous and per-item. Adherence to the supplied context can be measured separately from correctness~\cite{es2023ragas,adlakha2024evaluating,zhou2023contextfaithful}; metric-first evaluations formalise assumptions~\cite{min2023factscore} and test shortcut robustness~\cite{gao2023alce}. FActScore's source precision and Adlakha's token-overlap K-Precision resemble our exposure count, and ALCE rejects ``copy the passage'' as a shortcut where we use it as the reference point. Contamination audits~\cite{sainz2023contamination,golchin2024timetravel} assess input-visible score post hoc; we define deliberate exposure as the reference a priori. The closest ancestry is the extractive-oracle and lead-$n$ baselines of neural summarisation~\cite{see2017pointergenerator,narayan2018xsum}, which likewise ask what a copy of the input already achieves.

\paragraph{Nearest neighbours, and how we differ.}
Closest in method is work that causally manipulates the gold answer to test its effect on retrieval-augmented gains~\cite{answerpresence2026}; that work is withdrawn, and we adopt only its paired-intervention method. Leakage-free benchmarks instead extract reasoning graphs from question--context pairs and apply type-constrained entity replacement with leakage filters~\cite{seedrg2026}, during corpus construction rather than as a per-item measurement. RAGChecker decomposes RAG quality into claim-level retrieval and generation scores through LLM extraction and grading~\cite{ragchecker2024}; our context utilisation is a deterministic, surface-matched version of its metric and we claim no originality for the decomposition. MIRAGE measures benchmark-level adaptability against oracle context across retriever--model pairings~\cite{mirage2025}. Judged arms use a cross-family judge to limit self-preference~\cite{zheng2023judging,panickssery2024selfpreference}.

\paragraph{Ground claimed, stated narrowly.}
Against the three closest instruments we claim less than a new construct: RAGChecker's context utilisation already performs the exposure/delivery decomposition, via LLM claim extraction rather than a deterministic matcher~\cite{ragchecker2024}; the reader--context diagnostic already reports graded, judge-free answer presence in packed context, validated by intervention~\cite{readercontext2026}, and is the closest comparator overall; and Sufficient Context already shows that context-derivable-answer sufficiency is a stronger and better-validated criterion than raw answer-string presence~\cite{sufficientcontext2025}. Our contribution is a cheap, deterministic, judge-free diagnostic and a reporting convention: four-count accounting with a signed copy-baseline scalar, a copy-as-answer invariant, a seed-disjoint placebo, and a stratified audit that prices what the matcher gives up, together with the failure modes we document and the decisions they change. We claim neither priority over any one component nor a general boundary between copying and reasoning.

\section{Curated-Corpus Setting}\label{sec:setting}
This node is a served binary\footnote{The served node \url{https://github.com/DreamLab-AI/loom} (code under the repository licence; corpus terms per the sibling knowledgeGraph repository).} operating over a corpus produced by a separate, CI-gated pipeline\footnote{The canonical builder \url{https://github.com/DreamLab-AI/knowledgeGraph} (published at \url{https://narrativegoldmine.com}) runs \texttt{pytest pipeline/tests} and \texttt{pipeline.validate} before publishing each generation.} and accelerated by an in-process HNSW index\footnote{RuVector: \url{https://github.com/ruvnet/ruvector}.}.

\subsection{Ecosystem in brief}\label{sec:ecosystem}
Knowledge enters through the \emph{knowledgeGraph} repository, which compiles markdown pages into a pure-TBox OWL~2 ontology with no individuals and publishes it as versioned generations. \textbf{The evaluated snapshot} is the scaffold index generated on 2026-08-15, with 8{,}146 classes. The build reported 8{,}138 pages and a reasoned closure of 282{,}492 triples for that generation; those are the counts used throughout, and every number in this paper refers to that snapshot.

\paragraph{Provenance and governance.} The corpus was developed to support VisionFlow and the author's research. Its content is almost entirely LLM-derived, produced under the author's direction. Third-party extraction output enters through a preview-first import that refuses overwrites and emits review candidates instead of publishing machine output~\cite{ontocast}. The documented governance has three parts: CI consistency tests together with a Whelk OWL~2~EL reasoner gate~\cite{whelk}, which rejects logical contradictions before a generation is published; OKF and vocabulary validation in CI; and a propose-then-human-merge path for admitting changes. These gates check consistency and entailment, not factual accuracy, and no systematic human review of the content is documented, so nothing here should be read as per-assertion human verification. ``Curated'' in this paper means selected, structured and maintained under the author's direction. The corpus is public at \url{https://narrativegoldmine.com}. Public availability makes it inspectable but does not establish that any evaluated model saw it in training; synthetic authorship establishes neither its novelty nor its reliability. After the snapshot the corpus was migrated from Logseq to an Obsidian vault (\emph{visionGraph}, 2026-09-02 to 2026-09-23), and it is larger today (8{,}446 pages on 2026-09-28).

\emph{Loom}, the measured node, mirrors and serves a generation. It combines lexical retrieval over class titles, confidence-gated injection of per-IRI markdown blocks, read-only SPARQL over the closure and delegation to a model through an OpenAI-compatible fa\c{c}ade, locally Qwen3.8-27B on two workstation GPUs. The served node and that local model are the subject of the production study, the negative controls and the podcast case study (\S\ref{sec:live}, \S\ref{sec:controls}, \S\ref{sec:casestudy}); the ten-model sweep of \S\ref{sec:tenmodel} instead runs the same frozen scaffold engine offline against ten API models. One of them, Qwen-2.5-72B, is a different model from the local 27B.

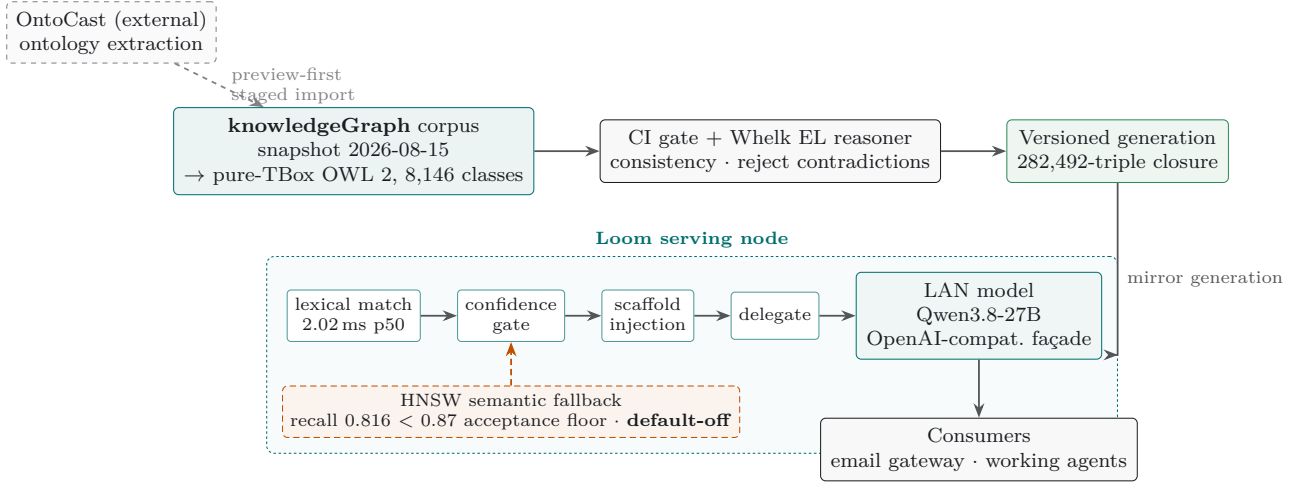
\begin{figure*}[t]\centering
\resizebox{\textwidth}{!}{%
\begin{tikzpicture}[
  font=\footnotesize,
  >={Stealth[length=2mm]},
  block/.style={draw=ink,fill=ink!3,rounded corners=2pt,align=center,inner sep=4pt,minimum height=8mm},
  corpus/.style={draw=teal0,fill=teal0!8,rounded corners=2pt,align=center,inner sep=4pt,minimum height=8mm},
  gen/.style={draw=softgreen,fill=softgreen!8,rounded corners=2pt,align=center,inner sep=4pt,minimum height=8mm},
  ext/.style={draw=ink!55,dash pattern=on 2pt off 2pt,fill=ink!2,rounded corners=2pt,align=center,inner sep=3.5pt},
  stage/.style={draw=teal0!70,fill=white,rounded corners=1.5pt,align=center,inner sep=3pt,minimum height=6.5mm,font=\scriptsize},
  offgate/.style={draw=burnt,dash pattern=on 2.5pt off 1.5pt,fill=burnt!7,rounded corners=2pt,align=center,inner sep=3pt,font=\scriptsize},
  flow/.style={->,ink!75,line width=0.7pt},
  offflow/.style={->,burnt,dash pattern=on 2.5pt off 1.5pt,line width=0.7pt},
]
% ---- ingestion row ----
\node[corpus] (kg) {\textbf{knowledgeGraph} corpus\\snapshot 2026-08-15\\$\rightarrow$ pure-TBox OWL~2, 8{,}146 classes};
\node[ext,above left=6mm and -6mm of kg] (ontocast) {OntoCast (external)\\ontology extraction};
\node[block,right=9mm of kg] (ci) {CI gate $+$ Whelk EL reasoner\\consistency $\cdot$ reject contradictions};
\node[gen,right=9mm of ci] (gen) {Versioned generation\\282{,}492-triple closure};
% ontocast -> kg via preview import
\draw[flow,ink!55,dash pattern=on 2pt off 2pt] (ontocast) -- node[right=1pt,font=\scriptsize,align=left]{preview-first\\staged import} (kg);
\draw[flow] (kg) -- (ci);
\draw[flow] (ci) -- (gen);
% ---- Loom serving node (below) ----
\node[stage,below=13mm of kg]        (match) {lexical match\\2.02\,ms p50};
\node[stage,right=5mm of match]      (gate)  {confidence\\gate};
\node[stage,right=5mm of gate]       (scaf)  {scaffold\\injection};
\node[stage,right=5mm of scaf]       (deleg) {delegate};
\node[block,right=5mm of deleg,fill=teal0!6,draw=teal0] (model) {LAN model\\Qwen3.8-27B\\OpenAI-compat.\ fa\c{c}ade};
\draw[flow] (match) -- (gate);
\draw[flow] (gate) -- (scaf);
\draw[flow] (scaf) -- (deleg);
\draw[flow] (deleg) -- (model);
% HNSW fallback (default-off)
\node[offgate,below=6mm of gate] (hnsw) {HNSW semantic fallback\\recall 0.816 $<$ 0.87 acceptance floor $\cdot$ \textbf{default-off}};
\draw[offflow] (hnsw) -- (gate);
% node grouping box
\begin{scope}[on background layer]
\node[draw=teal0,dash pattern=on 1pt off 1pt,rounded corners=3pt,fill=teal0!3,
      fit=(match)(model)(hnsw),inner sep=6pt,label={[teal0,font=\scriptsize\bfseries]above:Loom serving node}] (loombox) {};
\end{scope}
% generation feeds the loom node (mirror)
\draw[flow] (gen.south) |- (loombox.east) node[pos=0.28,right,font=\scriptsize]{mirror generation};
% consumers
\node[block,below=8mm of model,align=center] (cons) {Consumers\\email gateway $\cdot$ working agents};
\draw[flow] (model) -- (cons);
\end{tikzpicture}}
\caption{The data path around the measured node. Knowledge enters at the \emph{knowledgeGraph} corpus;
the evaluated snapshot is the scaffold index of 2026-08-15 (8{,}146 classes; 8{,}138 pages and a
282{,}492-triple reasoned closure as reported by that build). The corpus is public and almost entirely
LLM-derived under the author's direction, which establishes neither that any evaluated model saw it in
training nor that its content is novel or reliable. Output from OntoCast, an independent third-party
extractor, enters only through a preview-first staged import that emits review candidates. The CI gate and
the Whelk OWL~2~EL reasoner check consistency and entailment and reject contradictions; they do not check
factual accuracy. The \emph{Loom} node mirrors a generation and serves it through a model-free pipeline
(lexical match at 2.02\,ms, confidence gate, scaffold injection) before delegating to whatever LAN model sits
behind its OpenAI-compatible fa\c{c}ade (Qwen3.8-27B here). The HNSW semantic fallback is drawn dashed
because it ships default-off: its measured recall (0.816) is below 0.87, a recorded implementation
acceptance floor with no recorded derivation, and the per-query trigger, when enabled, is keyed to
lexical absence (\S\ref{sec:ecosystem}, \S\ref{sec:paraphrase}).}
\label{fig:ecosystem}
\end{figure*}

Loom is one component of a larger stack, and two other components matter here because the write-path results are reproducible only against them. Ingestion and the ledger writer of \S\ref{sec:lifecycle} run in the \emph{VisionFlow} agent environment, where the podcast extraction of \S\ref{sec:casestudy} executes; admission to the served corpus runs through the graph's governed propose/approve path, the ontology-bridge component of the same stack. Figure~\ref{fig:ecosystem} draws the read path around the measured node and \S\ref{sec:lifecycle} the write path. The instrument is corpus-general; the write-path study is checkable only where the write path lives.

Five deployment requirements drive these choices: a corpus the operator controls, auditability at single-entity granularity, accelerators that may locate the served markdown unit but never stand in for it, LAN confinement with the model delegated by URL, and confidence-gated injection. Together they call for high recall where the corpus is the intended source, attribution metadata on every answer (whose accuracy is unmeasured here, Table~\ref{tab:instruments}), and no regression on general questions the model already answers. ``Private'' in this stack's documentation denotes that LAN-confined deployment pattern, not a secret corpus.

\section{Ontology Scaffold}\label{sec:system}
Loom is a single static Rust binary built as a hexagonal workspace: a domain core defining the port traits, one adapter per backend, a scaffold-policy module and a thin HTTP façade. Only the domain core mints a served unit, so no adapter can return its own row, triple or vector as the payload, and golden fixtures pin retrieval, gating and scaffold serialisation byte-for-byte to a reference implementation.

\paragraph{Canonical unit and the corpus.} The unit of service and review is a markdown-with-ontology block scoped to one IRI: curated research prose preceded by typed relations (\texttt{subClassOf}, \texttt{requires}, \texttt{enables}, \texttt{uses}, \texttt{relatedTo}, \texttt{contrastsWith}). The corpus is a single sha-addressable \emph{generation}, here \textbf{8{,}146 concept classes} and \textbf{282{,}492 triples} in the Whelk-reasoned closure, mirrored atomically so the node serves a known, mixed-build-free version. Where other approaches reduce knowledge to opaque community summaries or GNN-encoded subgraphs~\cite{edge2024graphrag,he2024gretriever}, this system keeps reviewable markdown as the durable asset and lets the accelerators only locate, rank or attest to it~\cite{garcez2023neurosymbolic}.

\begin{table*}[t]\centering\small
\caption{Request pipeline of the Loom serving node. Retrieval, gating and scaffold serialisation
need no model; only the final delegation calls one. The single injection authority (match $+$ gate) is the sole path by
which any candidate reaches the model, whether lexical or, when enabled, semantic. Retrieval was
lexical only in every experiment reported: the semantic fallback was off throughout.}
\label{tab:pipeline}
\renewcommand{\arraystretch}{1.15}
\begin{tabularx}{\textwidth}{@{}>{\raggedright\arraybackslash}p{2.7cm} >{\raggedright\arraybackslash}X >{\centering\arraybackslash}p{2.2cm}@{}}
\toprule
Stage & Mechanism & Model? \\
\midrule
Match & Lexical inverted index over 8{,}146 class titles; 2.02\,ms p50 $\to$ seed classes & no \\
Gate & Single injection authority: budget scaled to retrieval score, skip below threshold & no \\
Scaffold & Budget-clamped serialisation of the matched units' markdown-with-ontology & no \\
SPARQL / search & Read-only, prologue-aware LIMIT-clamped oxigraph over the reasoned closure & no \\
Semantic fallback & In-process HNSW over 384-d embeddings; fires on lexical miss $\to$ candidate seeds into the same gate & \emph{default-off} \\
Delegate & Merge scaffold into the system message (last user turn is the retrieval query); call the configured backend URL & yes \\
\bottomrule
\end{tabularx}
\end{table*}

Together, the lexical matcher and confidence gate form the exclusive injection authority: one policy scales the scaffold budget to the retrieval score and injects nothing below a defined threshold (\S\ref{sec:ood}). Generation is delegated through the model-swap seam to the configured backend URL, currently Qwen3.8-27B, and responses carry model identity, so \S\ref{sec:method} can vary the model while holding grounding constant, or vary the serving path while holding the model constant. The \textbf{2.02\,ms} figure covers the model-free lexical match alone; generation dominates a complete grounded answer, at $\approx90$--$170$\,s median end-to-end across the Study~2 sets (\S\ref{sec:live}).

\section{The Copy Ceiling}\label{sec:ceiling}
Exposure accounting is the paper's central contribution and the copy ceiling its reportable scalar. Following the metric-first precedent~\cite{min2023factscore,gao2023alce}, this section gives a formal definition of both, an explicit list of assumptions and a robustness design. The order matters: readers will quote the scalar, but the accounting is what survives a change of implementation.

\subsection{Definition and assumptions}
Fix a question $q_i$ with gold target set $G_i$ (the \{slug, title\} pairs the graph asserts as its answer), the exact scaffold text $S_i$ shown to the model, and answer $a_i$. Let $\mathrm{m}(t,X)\in\{0,1\}$ be the deterministic surface matcher, equal to one when the normalised title of gold item $t$, or at least $80\%$ of its length-$\geq4$ words, appears in $X$. Define the exposed count and the per-item \textbf{copy ceiling}, an input-exposure reference and \emph{not} an upper bound on admissible answers,
\[
  e_i=\sum_{t\in G_i}\mathrm{m}(t,S_i),\qquad
  c_i=\frac{e_i}{|G_i|},
\]
and the grounded recall $r_i=\bigl(\sum_{t\in G_i}\mathrm{m}(t,a_i)\bigr)/|G_i|$. A no-op extractor returning $S_i$ verbatim scores exactly $c_i$. Read the name as ``the recall a copy already achieves'', never as a limit a good answer may not pass: a model can and sometimes does exceed it, and a perfect answer may sit exactly on it. Exposure is computed over the full visible input, the injected scaffold plus the question text. On the frozen sets the question text adds no exposure beyond the scaffold on any of the 510 questions. For the raw arm (question text only) exposure is small but non-zero: 35 of 1{,}136 gold items appear in the bare question, a mean raw-arm ceiling of 0.037.

\paragraph{Scoring symmetry, stated and checked.} One question template (T-COMMON, 45 of 510 questions) accepts \emph{any} of several valid targets, and the released implementation computes the ceiling as the exposed \emph{fraction} for these questions while the answer scorer credits any match. On the frozen sets this asymmetry is numerically inert, because every multi-target question's alternatives are exposed all-or-nothing. The symmetric scorer reproduces every reported gain to four decimal places, and the executable invariant (the concatenated visible input, scored as an answer, equals the reported per-item ceiling) holds on all 510 questions under both definitions and runs as a hard check in the released script. The invariant does not resolve the gold-set issue of \S\ref{sec:method}: a question-level check cannot tell a required component from an interchangeable alternative.

\paragraph{Assumptions (when the instrument applies).}
\begin{enumerate}[label=(A\arabic*),topsep=2pt,itemsep=1pt,leftmargin=2.4em]
\item \textbf{Deliberate exposure.} Questions, gold and scaffold derive from one graph, so gold is deliberately present in the injected text. The ceiling measures this exposure; it is not a leak.
\item \textbf{Shared matcher.} $c_i$ and $r_i$ use the same matcher. The matcher is a surface proxy: it under-counts paraphrase, and it can also over-count (its exact-title path is a substring test, and its long-word threshold ignores word order, relation roles and negation, so naming a correct entity in a wrong or negated relation still scores). Absolute recall is therefore neither a strict lower nor a strict upper bound on semantic correctness. Because one matcher scores both sides, the two scores share their error sources and some of that error may cancel in the paired difference $r_i-c_i$. How much cancels is an empirical question, not a guaranteed order of robustness, since error rates can differ between structured scaffold text and generated prose. \S\ref{sec:audit} measures the matcher's error rates on both sides of the accounting, and \S\ref{sec:rescore} re-scores under a paraphrase-tolerant matcher.
\item \textbf{Per-item, model-free ceiling.} $c_i$ depends only on $(q_i,S_i)$ and is identical across models given a fixed scaffold, which is what allows comparison across a model sweep.
\item \textbf{An appropriate gold set and matching criterion.} The ceiling requires a gold target set that is the right answer to the question and a matcher whose criterion suits that set. Gold and context deriving from one source is not itself a requirement; it is the case in which the exposure confound is most acute, and so the case in which the reference is most needed. Where gold is independently authored (the out-of-domain arm), $c_i\approx0$ and judged quality carries the axis.
\end{enumerate}

\paragraph{A recall reference point, not an endorsement.} Although the no-op extractor scores $c_i$, it is not an admissible answer: returning the entire scaffold lacks the required precision, conciseness and relevance. The ceiling is a recall reference for delivery, not a recommendation to reproduce the context.

\subsection{Exposure accounting: the counts are primary}\label{sec:accounting}
For each gold item $t\in G_i$ write $x_t=\mathrm{m}(t,S_i)$ for \emph{exposed} and $y_t=\mathrm{m}(t,a_i)$ for \emph{recovered}. Pooling over all gold items of a run gives four counts,
\[
\begin{aligned}
  n_{11}&=\#\{x_t{=}1,\,y_t{=}1\}, & n_{10}&=\#\{x_t{=}1,\,y_t{=}0\},\\
  n_{01}&=\#\{x_t{=}0,\,y_t{=}1\}, & n_{00}&=\#\{x_t{=}0,\,y_t{=}0\},
\end{aligned}
\]
with $N=n_{11}+n_{10}+n_{01}+n_{00}$. These four counts determine the micro-averaged exposure and recovery metrics listed in this subsection, and only those. Macro-averaged question scores, question-level uncertainty and judged outcomes need the per-question records, since different allocations of the same four counts across questions give different macro scores. The pooled ceiling is the exposed fraction $(n_{11}+n_{10})/N$, pooled recall is $(n_{11}+n_{01})/N$, and their difference, the pooled gain over copy, is
\[
  \bar g=\frac{n_{01}-n_{10}}{N},
\]
the net of two cells. Context utilisation, $n_{11}/(n_{11}+n_{10})$, is the fraction of exposed gold that survives delivery; beyond-exposure recovery, $n_{01}/(n_{01}+n_{00})$, is the fraction of unexposed gold the model supplies itself. The headline figures in \S\ref{sec:results} average per-question ratios rather than pooling items, so they differ from these identities in the second decimal; the released decomposition script reports both.

Two properties make the counts the right object to report. First, they are diagnostic where the scalar is not: a gain of zero is consistent with $n_{01}=n_{10}=0$ and with $n_{01}=n_{10}=200$, a model dropping as much exposed gold as it invents, and only the counts tell those apart. Second, each $x_t$ depends on the matcher and fixed text alone, with no ranking, threshold or tuning. Where a task has no natural no-op extractor a baseline must instead be \emph{constructed} from a ranker, which is a different instrument and not the quantity defined here (\S\ref{sec:companion}). The ceiling is the number to quote; the counts are the number to trust.

\subsection{Gain over copy}
We define the signed \textbf{gain over copy} as $g_i=r_i-c_i$, with headline figure $\bar g=\bar r-\bar c$. It re-centres grounded recall on the exposure reference; pooled over items it is $(n_{01}-n_{10})/N$ (\S\ref{sec:accounting}).
\begin{itemize}[topsep=2pt,itemsep=1pt,leftmargin=1.4em]
\item $g\approx0$: the model recovers about as much as the context exposes. This is a \emph{net}, not a statement of fidelity: it holds when $n_{01}=n_{10}=0$, and equally when the model drops 200 exposed items and supplies 200 unexposed ones. Only the counts separate the two, which is why they, and not the scalar, carry the findings.
\item $g<0$, \emph{net omission}: answers omit more exposed gold names than they supply unexposed ones. The decomposition of \S\ref{sec:tenmodel} measures the two cells directly, and here the deficit is almost entirely exposed-but-omitted gold ($n_{10}$), with $n_{01}$ negligible. Omission is an outcome, not a mechanism: selection of the most relevant names, reformulation the matcher does not catch, choosing one accepted alternative, optional gold, genuine omission and truncation can all contribute, and this study does not isolate them. The magnitude places models on an exposure-normalised scale, which raw uplift cannot provide because it conflates exposure with delivery.
\item $g>0$: the model recovers gold beyond what the copy baseline finds. The sign does not identify its source: parametric memory, guessing, paraphrase that the matcher catches in the answer but not in the context, and matcher asymmetry all produce it, and this instrument cannot separate any of them from reasoning.
\end{itemize}
A result near the copy baseline shows that the measured recall is achievable from exposure alone. An explanation of the score that involves no inference is then always available, but it is not the only one: the metric does not observe how a model selected the names it returned, so it constrains what the evaluation can attribute, not what the model did.

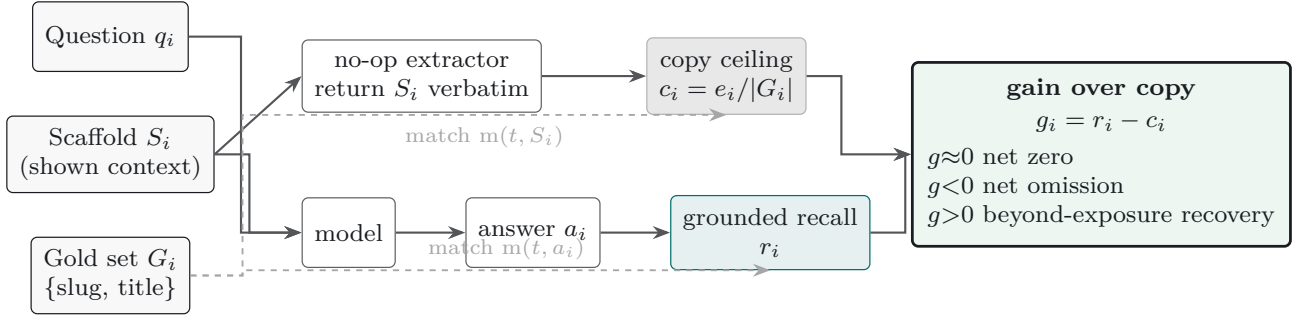
\begin{figure*}[t]\centering
\resizebox{\textwidth}{!}{%
\begin{tikzpicture}[
  font=\footnotesize,
  >={Stealth[length=2mm]},
  inp/.style={draw=ink,fill=ink!3,rounded corners=2pt,align=center,inner sep=4pt,minimum height=8mm},
  proc/.style={draw=ink!70,fill=white,rounded corners=2pt,align=center,inner sep=4pt,minimum height=8mm},
  ceil/.style={draw=gray!60,fill=gray!18,rounded corners=2pt,align=center,inner sep=4pt,minimum height=8mm},
  rec/.style={draw=teal0,fill=teal0!10,rounded corners=2pt,align=center,inner sep=4pt,minimum height=8mm},
  gain/.style={draw=ink,line width=0.9pt,fill=softgreen!8,rounded corners=2pt,align=center,inner sep=5pt},
  flow/.style={->,ink!75,line width=0.7pt},
  ref/.style={->,gray!70,dash pattern=on 2pt off 2pt,line width=0.6pt},
]
% inputs
\node[inp] (scaf) {Scaffold $S_i$\\(shown context)};
\node[inp,above=5mm of scaf] (q) {Question $q_i$};
\node[inp,below=5mm of scaf] (gold) {Gold set $G_i$\\\{slug, title\}};
% top path: no-op extractor -> copy ceiling
\node[proc,right=10mm of scaf,yshift=9mm] (noop) {no-op extractor\\return $S_i$ verbatim};
\node[ceil,right=12mm of noop] (ci) {copy ceiling\\$c_i=e_i/|G_i|$};
% bottom path: model -> answer -> recall
\node[proc,right=10mm of scaf,yshift=-9mm] (model) {model};
\node[proc,right=8mm of model] (ans) {answer $a_i$};
\node[rec,right=8mm of ans] (ri) {grounded recall\\$r_i$};
% flows
\draw[flow] (scaf.east) -- (noop.west);
\draw[flow] (q.east) -- ++(6mm,0) |- (model.west);
\draw[flow] (scaf.east) -- ++(4mm,0) |- (model.west);
\draw[flow] (noop) -- (ci);
\draw[flow] (model) -- (ans);
\draw[flow] (ans) -- (ri);
% gold is the shared matcher reference for both scores
\draw[ref] (gold.east) -- ++(6mm,0) |- (ci.south) node[pos=0.75,below,font=\scriptsize]{match $\mathrm{m}(t,S_i)$};
\draw[ref] (gold.east) -- ++(6mm,0) |- (ri.south) node[pos=0.75,above,font=\scriptsize]{match $\mathrm{m}(t,a_i)$};
% gain over copy
\node[gain,right=12mm of ci,yshift=-9mm] (g) {\textbf{gain over copy}\\[1pt]$g_i=r_i-c_i$\\[3pt]
  \begin{tabular}{@{}l@{\ }l@{}}
    $g{\approx}0$ & net zero\\
    $g{<}0$ & net omission\\
    $g{>}0$ & beyond-exposure recovery\\
  \end{tabular}};
\draw[flow] (ci.east) -- ++(4mm,0) |- (g.west);
\draw[flow] (ri.east) -- ++(4mm,0) |- (g.west);
\end{tikzpicture}}
\caption{The copy baseline (copy ceiling) and the signed gain over copy on one panel. The copy
comparison tests what a recall score establishes; it does not test whether reasoning occurred: when an
answer name is already in the shown context, a reasoned answer and a verbatim copy score the same, and
the copy is a recall reference, not an upper bound or a better answer. The same shown context $S_i$ feeds
two extractors: a no-op extractor that returns $S_i$ verbatim, whose recall against the gold set $G_i$
is the per-item \emph{copy ceiling} $c_i$; and the model, whose answer $a_i$ scores the grounded recall
$r_i$. Both scores use the identical surface matcher $\mathrm{m}(t,\cdot)$ with $G_i$ as reference
(dashed), so their difference, the \emph{gain over copy} $g_i=r_i-c_i$, shares its error sources and
some of that error may cancel; how much is an empirical question, since error rates can differ between
scaffold text and generated prose (\S\ref{sec:audit}). The sign reads as a net, not as a verdict on
fidelity: $g\approx0$ means the model recovered about as much as the context exposed, which holds both
when nothing was dropped or added and when omissions and recoveries cancel; $g<0$ is net loss of
exposed gold names, whatever its cause; and $g>0$ is recovery beyond what the exposure check finds, without identifying its
source (\S\ref{sec:ceiling}).}
\label{fig:ceiling}
\end{figure*}

\paragraph{What zero gain does and does not say.} The accounting does not test for the presence or absence of reasoning, and no result in this paper should be read as if it did. In the introduction's example (Project~A uses Module~B; Module~B requires Component~C), choosing Component~C requires following two relations, yet its name is already a string in the context, so exposure is 1, a perfect answer scores 1, and gain over copy is exactly zero. Zero gain is therefore compatible with successful reasoning over exposed names, and negative gain with reasoning that sometimes fails; positive gain is compatible with memorisation, guessing or matcher asymmetry. The example generalises: wherever a curated corpus is served one entity-scoped block at a time, the names an answer needs are usually already on the page, and a recall metric over those names cannot see which route the model took to them. The defensible statement is the narrow one, \emph{this recall score does not establish reasoning beyond answer-name exposure}, and establishing whether a model composes exposed structure needs a second instrument that varies the route while holding the answer name exposed. This paper does not run that instrument; it keeps the two questions apart. \S\ref{sec:audit} shows empirically where the exposure instrument stops: roughly one credited item in sixty names the gold target without asserting the relation the question asked for. Figure~\ref{fig:ceiling} draws the two paths being compared.

\subsection{Negative controls: what would move the gain}\label{sec:ceiling-controls}
A positive delivery result might reflect lexical overlap rather than knowledge transfer. Five arms test this by holding the model and question fixed and perturbing only the injected block (defined here; results in \S\ref{sec:controls}):
\begin{description}[topsep=2pt,itemsep=1pt,leftmargin=0pt,font=\normalfont\itshape]
\item[true] the block served by the node, injected into the bare model without the live path's authority preamble, which isolates the block from the rest of the serving path.
\item[shuffled] the same block with body sentences shuffled. Whole-sentence shuffling preserves every within-sentence fact, so this arm tests sensitivity to sentence \emph{order} only.
\item[masked] the same block with every seed-class title replaced by \texttt{Entity-$k$}. Answer-bearing \emph{target} names remain, so this probes dependence on seed naming, not on answer-name matching.
\item[irrelevant] the well-formed, on-corpus scaffold of a \emph{different} question, using a fixed derangement over engaged items. Seed-IRI disjointness is a construction check: it prevents donor and target sharing seed classes, but does not guarantee that the donor block exposes none of the target's gold, ancestors or synonyms, and the five-arm rerun measures that exposure (\S\ref{sec:controls}).
\item[fluent noise] sections drawn at random from the same scaffold index, restricted to classes disjoint from the target's seeds, their one-hop relation targets and their named ancestors, rendered by the live path's own section renderer and packed to the true block's token count. It keeps register and length but removes the coherence a donor scaffold still has as an answer to \emph{some} question.
\end{description}
Each arm answers a different question. \emph{Shuffled} and \emph{masked} ask whether sentence order and seed naming carry any of the effect. \emph{Irrelevant} and \emph{fluent noise} ask whether the specific retrieved content matters beyond the presence of a well-formed, on-corpus block. Neither is a zero-content floor: both are drawn from the same corpus, and a lexical matcher finds some of a question's gold names in them by coincidence, which is why their exposure is measured rather than assumed. Together they test whether the observed lift is content-specific, the direct analogue of a shortcut-robustness check for a context-adherence metric~\cite{gao2023alce}.

\subsection{Relation to prior instruments}\label{sec:ceiling-prior}
Three recent instruments ask a related question (did the shown context deliver the answer?), and \S\ref{sec:related} concedes precedent on each component. This subsection places the copy ceiling among them. RAGChecker's context utilisation performs the same decomposition through LLM claim extraction and entailment checking~\cite{ragchecker2024}; restricting it to a deterministic surface matcher buys zero judge cost and exact reproducibility, and costs semantic discrimination. The reader--context diagnostic is the nearest neighbour, scoring graded, judge-free answer presence in packed context~\cite{readercontext2026}; the distinction is accounting and apparatus, not the underlying presence idea. \emph{Sufficient Context} asks a strictly stronger question, whether \emph{an} answer is derivable from the context at all, and its ablation shows that criterion beating a ground-truth-string-in-context baseline, $93\%$ against $81\%$ agreement~\cite{sufficientcontext2025}; our exposure ceiling is deliberately that weaker baseline. The idea of an input-side reference is older still: the extractive-oracle tradition in summarisation~\cite{see2017pointergenerator,narayan2018xsum,xu2019extractive} makes a copy of the input the natural reference for any recall-style metric, and an information-bottleneck view compresses retrieved context against the answer for a related purpose~\cite{zhu2024ibrag}. The trade-off is deliberate, and it is why \S\ref{sec:audit} exists: removing semantic judgement removes semantic discrimination, so the ceiling is read as a delivery-recall reference and the audit prices what that reading omits.

\section{Semantic Re-Score: Is the Ceiling a Lexical Artefact?}\label{sec:rescore}
Scoring with a paraphrase-tolerant matcher leaves the negative gain in place. Because the matcher is lexical, a sceptic
can ask whether the negative gain is an artefact of strict matching, with answers paraphrasing gold
that the ceiling, computed on the same matcher, happens to catch. We re-score every saved sweep row with a paraphrase-tolerant embedding matcher, re-scoring the
copy ceiling under the identical matcher so the comparison stays symmetric. The matcher embeds
candidate spans (word 1--6-grams within delimiter-bounded chunks, plus each chunk and the whole text)
with \texttt{bge-small-en-v1.5} and scores a gold item as recovered when the maximum cosine against
its title meets a threshold; we sweep thresholds $0.80/0.85/0.90$ and use the paper's seeded
10{,}000-resample paired bootstrap. The copy contexts here are \emph{reconstructed}, not shown to be byte-identical to
those of the original sweep, because the released rows store each question's exposure count rather
than its context string. They are regenerated with the frozen scaffold engine and admitted only under an
equivalence gate, which requires every question's recomputed exposed-gold count to equal the count
stored on its sweep row and the mean lexical ceiling to agree to four decimal places ($0.9645$).
Two further quantities match the original run: the regenerated set yields 778 gold titles, the
figure that run recorded, and a 126-item sanity sample of lexical hits agrees with semantic scoring
at 100\% at every threshold. The gate checks equivalence of exposure and of lexical scoring only. Two texts with equal lexical
exposure may still be worded differently, and embedding-match behaviour could differ between them,
so this is a re-score on reconstructed contexts, not on the original bytes.

Level moves with the threshold; sign does not. At $0.80$ and $0.85$ semantic scoring
raises both recall and ceiling, the ceiling reaching $0.980$ and $0.968$ against $0.9645$ lexical; at
the strictest threshold, $0.90$, both fall below their lexical values ($0.937$ ceiling). In all
\textbf{thirty} model~$\times$~threshold combinations the gain over copy stays negative with the
95\% CI excluding zero. Per-model magnitudes move slightly, the largest shift being Qwen-2.5-72B from
$-0.050$ lexical to $-0.034$ at $0.80$, but no model's interval crosses zero. Pooled beyond-exposure
recovery stays at 3 of $11{,}360$ at both $0.80$ and $0.85$, where the pooled decomposition is
$n_{11}=10{,}001$, $n_{10}=669$, $n_{01}=3$, $n_{00}=687$; only at $0.90$, where the matcher has
become strict enough to lose exposure as well as recall, does it reach 20, and no single model
exceeds 4 at any threshold. The headline is therefore not an artefact of lexical matching.

Unlike the lexical instrument, this check depends on an embedding model and its span extraction; it
is a robustness probe on the sign of the gain, not a replacement scorer, and it speaks to lexical
strictness rather than to relational correctness, which is the subject of \S\ref{sec:audit}.

\section{Making the Copy Ceiling Runnable}\label{sec:runnable}
Needing no model, the control is cheap to adopt. To compute it on any corpus:
\begin{enumerate}[topsep=2pt,itemsep=1pt,leftmargin=1.6em]
\item \textbf{Inputs.} A list of questions, each with a gold target set (the \{slug, title\} pairs the source asserts as its answer) and the exact context string $S_i$ that would be shown to the model. The ceiling needs a gold target set appropriate to the question and a matching criterion suited to it (assumption~A4); shared provenance between gold and context is not required, though it is the case where the exposure confound bites hardest.
\item \textbf{Matcher.} A deterministic surface matcher $\mathrm{m}(t,X)$: one when the normalised title of gold item $t$, or $\geq80\%$ of its length-$\geq4$ words, appears in $X$. The reference implementation and the scaffold engine used here are released in the harness (\texttt{tools/paper/decompose\_exposure.py} and \texttt{ontology\_scaffold\_v1.py}); a paraphrase-tolerant embedding variant is a drop-in (\S\ref{sec:rescore}).
\item \textbf{Compute, no model call.} Per question, the copy ceiling is $c_i=\bigl(\sum_{t\in G_i}\mathrm{m}(t,S_i)\bigr)/|G_i|$: the fraction of gold already exposed in the context. The headline is the mean over questions. Grounded recall $r_i$ reuses the same matcher on the model's answer; gain over copy is $r_i-c_i$.
\end{enumerate}
\textbf{Reading the number.} A copy baseline near 1 ($0.964$ here) says the context already contains the answer strings, so retrieval supplies most of the name coverage. The reference then belongs \emph{beside} the quality and system baselines a generative component is judged on, not in place of them. The scalar does not rank components on answer quality: a component at zero or negative gain has not thereby been shown to be worse than one with positive gain, and a copy of the context is not an admissible answer, since nothing here scores it for concision, usability or whole-answer precision. A baseline near $0.5$ or below says the answer strings are largely missing. The first checks are then whether retrieval failed, whether the question was phrased outside the graph's vocabulary (\S\ref{sec:paraphrase}, where phrasing alone moves this corpus from $0.964$ to $0.328$), whether gold is incomplete and whether the matcher missed; only once those are ruled out does a low value bear on whether a task needs composition, and it never establishes that need on its own. In both directions the number says where to look first, not what the model is doing, and never that the required relation is present. The baseline prices none of user benefit, curation, governance, model substitution or reasoning; it is a diagnostic available once questions, gold and candidate contexts exist. A worked question, its gold and a sample scaffold block are in Appendix~\ref{sec:appendix-samples}.

\section{Vocabulary Mismatch: the Retrieval Boundary}\label{sec:paraphrase}
Lexical retrieval fails when a question is phrased outside the graph's title vocabulary, and the
node's default-off semantic fallback exists for that case. A paraphrase stress set tests how often
it arises and whether the fallback would catch it. Each of the 510 sweep questions is rewritten by \texttt{gpt-4.1} at temperature~0 under seed~42
to preserve meaning while excluding the seed class's title tokens, and each candidate passes a
three-part gate before entering the set: no gold title appears verbatim; no content token of the seed
class's title appears by exact or prefix match; and the asked relation is preserved, adjudicated by
the same model. Candidates are retried up to four times. \textbf{506 of 510} questions clear the gate
at a mean of $1.112$ attempts, four remain unresolvable, and the final set contains zero title-token
leaks by construction. The 61 rejected attempts divide into 33 that named the gold verbatim, 23 that
failed to preserve the relation and 5 that leaked a seed-title token.

Here the instrument is the copy ceiling itself, computed per arm with no generation: does the retrieved
block expose the gold? Two arms are compared. The \emph{lexical} arm queries the live production
fa\c{c}ade. The \emph{semantic} arm takes top-5 cosine seeds over all 8{,}146 class embeddings and
feeds them into the same frozen scaffold engine at the same budget. The fa\c{c}ade's own semantic
path is not reachable: \texttt{/health} reports it unready because the deployed index artefact
declares no embedding model and fails the contract check, so the semantic arm is necessarily a
harness-side reconstruction over the same corpus and the same embedding model.

\paragraph{Exposure collapses.} Mean ceiling falls from \textbf{0.964} on the original questions to
\textbf{0.328} on the paraphrases, with \textbf{64.4\%} of questions (326 of 506) below $0.5$. On the
original questions the live fa\c{c}ade and the frozen engine agree to four decimal places, confirming
that the deployed node retrieves what the frozen harness does.

\paragraph{Silent failure.} Partial word overlaps almost always engage some seed, so the
fallback's designed absence-keyed trigger (default off in deployment) would fire on only \textbf{2 of 506} questions ($0.40\%$), while 326 have
a ceiling below $0.5$. Returning something is not returning enough: a trigger keyed to finding nothing cannot detect finding the wrong thing, so the
fallback as designed is inert against the failure that occurs. This diagnoses the lexical retrieval path as tested, not semantic retrieval in general.

\paragraph{A real signal, wasted by the gate.} The semantic arm lifts the mean ceiling to $0.596$,
a recovery of $+0.269$ $[+0.227,+0.310]$ over the whole set and $\mathbf{+0.446}$
$[+0.398,+0.493]$ on precisely the 326 questions where lexical retrieval fails. The information
needed to rescue those questions is in the corpus and reachable by a path the shipped gate never
invokes.

\paragraph{Prescription, and its bounds.} The specific, supported prescription is that a hybrid
gate must be keyed to retrieval \emph{confidence} rather than to retrieval \emph{presence}: presence
is satisfied on $99.6\%$ of paraphrases while exposure is lost on $64.4\%$ of them. Whether semantic
seeds should then \emph{replace} lexical ones or be appended to them under a fixed token budget is
not measured here, but three considerations argue against wholesale replacement. On entity and
rare-token queries dense retrieval is the fragile arm and lexical matching holds
up~\cite{entityquestions2021}, so the safe form gates on query type or on dense-versus-lexical rank
disagreement and fuses the rankings~\cite{rrf2009}. Score-distribution predictors are weakest
precisely on semantic queries and cannot see a confident-but-wrong lexical seed~\cite{qppneural2023},
so the informative key is disagreement between the two retrievers rather than a single-ranking
confidence score. And the collapse reported here is a retrieval-\emph{exposure} measurement, whether
the block reaches the model at all, not the model's answer robustness: the paraphrase-robustness
literature reports single-digit accuracy drops~\cite{benchrobust2025} because it measures a different
construct downstream of successful retrieval, and the two figures are not comparable. Here the
ceiling does three jobs: it shows where exposure is lost, measures what a semantic path could
recover, and prices the design that would recover it.

\section{Method}\label{sec:method}
\paragraph{Design.} Both studies use within-model pairs: the \emph{same} model answers the \emph{same} question twice, which cancels between-question variance in difficulty. The sweep varies the \emph{model} with grounding fixed; the production study varies the \emph{serving path} with the model fixed.

\paragraph{Study 1: the ten-model sweep.} The knowledge graph supplies templated questions for every sufficiently-defined class: \textbf{T-REL} (requirements, uses, dependencies, enablers, and parts), \textbf{T-TAX} (immediate and ancestral parents), and \textbf{T-COMMON} (shared ancestors of a pair). Gold answers are the \{slug, title\} targets asserted by the graph. Seed~42 yields \textbf{510 questions} across 15 domains: 285 T-REL, 180 T-TAX, and 45 T-COMMON. Each is posed \emph{raw} and \emph{scaffold} (with a budget-clamped 1{,}500-token ontology extract prepended) to ten models from eight developer families (Google, Anthropic, OpenAI, Zhipu, DeepSeek, Meta, Alibaba, Mistral), accessed through their providers' APIs or a router, under one fixed configuration (\texttt{temperature}=0, \texttt{max\_tokens}=2048; \texttt{reasoning\_effort}=low only where thinking cannot be disabled, preventing mandatory passes from truncating answers). The lexical matcher from \S\ref{sec:ceiling} counts a gold item when its normalised title, or $\geq80\%$ of its long words, appears in the answer. The matcher is a surface proxy that can both over-count and under-count, so absolute recall is not a bound in either direction (A2); the signal is the lexical \emph{paired} delta for the same question, scorer and model. Truncation is detected per row: GLM-4.6 truncated on 174 rows, of which 5 were retried and the rest scored as received, making that row lower-confidence, and Gemini~2.5~Flash-Lite on 16.

\paragraph{What the scorer does and does not validate.} Two properties of the instrument bound every number the sweep reports. First, the matcher tests for the presence of a gold target's name, not that the answer asserts the right relation, that subject and object are the right way round, that the answer is not simultaneously denying the relationship, or that it is not an unusably broad list; an answer can therefore score fully while being relationally wrong. Using the same matcher on context and on answer does not guarantee that such errors cancel, because structured scaffold text and generated prose differ in length and syntax. \S\ref{sec:audit} measures this gap directly on a stratified sample of the sweep and reports both the rate and the mechanisms. Second, the gold sets are not uniformly conjunctive: T-COMMON (45 of 510 questions, of which 17 carry two or more acceptable targets) accepts \emph{any one} of several valid targets, while the item-level decomposition counts each target separately. A model can therefore give a fully acceptable answer and still accumulate exposed-but-omitted items, so item-level omission counts include some compliant omissions; \S\ref{sec:tenmodel} recounts them and \S\ref{sec:audit} checks that recount against judged assertions. Separating required components, interchangeable acceptable answers and optional additions in the gold sets remains to be done.

\paragraph{Study 2: the production-node paired study.} Study~2 holds the model constant at Qwen3.8-27B on the reference GPU host and varies only the serving path. The \emph{loom} arm calls the production Rust node's \texttt{/v1/chat/completions} endpoint, which retrieves, confidence-gates and injects the scaffold, returning telemetry (engagement, injected tokens, fusion path) in the response \texttt{loom} block. The \emph{raw} arm calls the same backend without a scaffold. Both use \texttt{temperature}=0 and \texttt{max\_tokens}=1536, below which the reasoning backend truncates output to empty. Three pre-generation frozen sets contain \textbf{117} questions: 24 \emph{arcane} (deep in-domain), 33 \emph{thin} (sparsely covered), and 60 \emph{general} (out-of-domain). The out-of-domain judged arm below also grades the general set, which keeps the judged arms comparable. The negative-control arms from \S\ref{sec:ceiling-controls} use the same node; their scaffolds come from the LLM-free \texttt{/loom/scaffold} endpoint, so each control presents the bare model, via a system message, with exactly the block bytes served to the loom arm.

\paragraph{Out-of-domain judged arm.} A five-model judged arm poses the 60-question \emph{general} set (adjacent, in-domain-general and off-domain subsets) to \emph{harnessed} and \emph{bare} configurations. Each pair is graded $0$--$5$ against frontier-authored key points, providing an independent out-of-domain instrument (\S\ref{sec:ood}).

\paragraph{Cross-family semantic judge.} Study~2 is scored by a reference-guided \textbf{0--5 rubric} judge rather than lexically, so that paraphrase is credited on both its in-domain and out-of-domain questions. The judge is \texttt{openai/gpt-4.1} via OpenRouter with \texttt{temperature}=0 and \texttt{max\_tokens}=200, applying the same fixed system-message rubric in every arm. It is from a different family from the Qwen candidate, which limits self-preference~\cite{zheng2023judging,panickssery2024selfpreference}; it is blind to each answer's arm and study origin, and grading can resume per $(\text{set},\text{id},\text{arm})$. References are graph-derived targets in-domain and frontier-authored key points out-of-domain.

\paragraph{Statistics.} Per set and pooled, the primary metric is the paired mean difference with a seeded \textbf{10{,}000-resample bootstrap} 95\% percentile interval~\cite{efron1979bootstrap,koehn2004significance}. Alongside it we report a two-sided \textbf{exact conditional signed-rank} test, which targets a rank summary rather than the mean and so complements it: zero differences are dropped (the reduced-sample procedure, not Pratt's), tied absolute differences receive average ranks, and the $p$-value comes from exact enumeration of the sign-randomisation distribution conditional on the observed ranks, which matters because one set has only two non-zero pairs. The paired effect size is the \textbf{matched-pairs rank-biserial correlation} $r=(T^{+}-T^{-})/(T^{+}+T^{-})$ over non-tied pairs. Set-level $p$-values receive a \textbf{Holm--Bonferroni} correction~\cite{dror2018hitchhiker}. For out-of-domain non-regression we use a conservative \emph{interval-inclusion} criterion in the spirit of equivalence testing~\cite{lakens2017equivalence}: non-regression is declared only when the 95\% bootstrap percentile interval lies wholly inside a pre-specified $\pm0.25$ margin, which is stricter than the 90\% interval conventional TOST would use. Because in-domain graph-derived questions cluster by class and domain~\cite{dror2018hitchhiker}, we report a domain-clustered block bootstrap alongside the naive interval and an intention-to-treat delta retaining questions where the gate did not engage~\cite{hollis1999itt}.

\section{Results}\label{sec:results}
Table~\ref{tab:instruments} lists the questions the studies answer, the instrument each uses and what each leaves open. The outcomes differ (lexical recall of gold target names, a model-judged relational label per item, a model-judged 0--5 answer-quality score, lexical term resolution and a model-judged page-improvement score), so they are reported separately and never pooled or compared cell for cell.

\begin{table*}[t]\centering\small
\caption{The questions this paper asks, the instrument that answers each, what was observed and
what remains unresolved. Each row uses its own outcome measure and population, so no number in one
row can be read on another row's scale. Grounding improved recall substantially (row~1), and a
separate production study found a model-judged quality benefit (row~3); the copy comparison does not
test whether reasoning occurred, and whether graph-structured grounding beats strong flat-text
retrieval is untested by any matched comparison (row~7). Rows~4 to~6 are operational studies; row~6
is a failure the recall measure could not have shown.}
\label{tab:instruments}
\renewcommand{\arraystretch}{1.15}
\begin{tabularx}{\textwidth}{@{}>{\raggedright\arraybackslash}p{2.9cm} >{\raggedright\arraybackslash}p{3.6cm} >{\raggedright\arraybackslash}X >{\raggedright\arraybackslash}p{4.3cm}@{}}
\toprule
Question & Instrument & Positive observation & What remains unresolved \\
\midrule
1. Are the answer names available? & Lexical exposure accounting and gain over copy; 510 questions, ten models (\S\ref{sec:tenmodel}) & Recall $0.265\to{\approx}0.92$ with grounding; copy baseline $0.964$; 3 of 11{,}360 observations recovered beyond exposure & Answer quality; how models reached the names; lexical only (paraphrase: \S\ref{sec:paraphrase}) \\
2. Are the relations asserted? & Stratified model-judged item audit, 423 sampled item observations (\S\ref{sec:audit}) & 233 of 240 exposed-and-credited observations assert the gold edge; estimated relational rate $0.905$ against lexical $0.931$ & Model judge, not human; the human sample is prepared, not annotated; whole-answer precision unscored \\
3. Does the deployed path improve answers? & Paired reference-based quality judgement, 0--5 scale, 114 pairs, second judge family; five-arm controls (\S\ref{sec:live}, \S\ref{sec:controls}) & $+0.27$ $[+0.11,+0.45]$; second family $+0.342$; every well-formed on-corpus block beats no context & Whether the specific content matters; one model; complete-case; no human judging \\
4. Does it hold beyond the corpus? & Judged $\Delta$ out of domain, five models and one general set (\S\ref{sec:ood}) & General set $+0.05$ $[0.00,+0.13]$ inside a $\pm0.25$ margin & Not established in the five-model worst case; instrument near its ceiling \\
5. Is extraction aligned with the corpus? & Term resolution, assertion count, latency; one podcast, ten episodes (\S\ref{sec:casestudy}) & Term resolution $0.17\to0.55$; $346\to173$\,s per episode; no speech-recognition artefact names in the grounded local arm & Lexical alignment, not assertion accuracy; one run \\
6. Does integration improve the page? & Blind before/after page judgement on a $-2..2$ improvement scale (\S\ref{sec:pagejudge}) & The judge caught degradation in every arm (means $-1.04$ to $-0.44$); integration was disabled & Whether the ledger redesign works is not measured \\
7. Does it beat strong flat-text retrieval? & None run; a matched design is proposed (\S\ref{sec:analysis}) & None & Unmeasured, including against direct graph query at equal budget \\
8. Is attribution accurate? & None run & Source-generation metadata is emitted on every answer & Accuracy of that metadata and precision against fabrication unmeasured \\
\bottomrule
\end{tabularx}
\end{table*}

\paragraph{Budget exhaustion and the complete-case sample (Study 2).} At \texttt{max\_tokens}=1536, 42 of 234 loom/raw completions, spanning 28 question pairs, exhausted the budget during reasoning and produced no output. Each affected pair was re-executed at 4096 in \emph{both} arms; three thin questions still came back empty in one arm (loom once, raw twice, so realised missingness was not symmetric) and were excluded pairwise, giving a complete-case $n=114$. The pooled contrast is therefore a mixed-budget estimand, with 25 re-run pairs at 4096 and 89 at 1536. The earlier four-arm control cohort of \S\ref{sec:controls} remained at 1536 and lost $36$--$50\%$ of its completions the same way; the five-arm rerun used 4096 tokens with one retry at 8192 and lost $1.8$--$5.4\%$.

\subsection{In-domain: recall against exposure}\label{sec:indomain}
Grounding raises recall far above the unaided level but leaves it just below the copy baseline. For Gemini~3.7~Flash, raw recall is 0.375 and scaffold recall 0.942, a paired delta of $\mathbf{+0.567}$ with naive and domain-clustered 95\% CIs of $[+0.532,+0.603]$ and $[+0.524,+0.611]$ across $n=510$ items, with no errors or truncated responses. Unaided recall is non-trivial across models ($0.151$--$0.375$); this study does not test what produces it, whether prior exposure to the public corpus, related public knowledge or cues in the question itself. The copy ceiling is \textbf{0.964}, placing recall $-0.022$ below the question-averaged exposed fraction. (That macro ceiling is not the item-pooled exposed fraction, $0.933$; the two are reported separately and are not interchangeable, \S\ref{sec:limits}.) Gemini's answers therefore name most gold targets the context exposed and omit some. Because recall does not measure additions beyond the gold, fabrication and attribution precision remain unmeasured (Table~\ref{tab:instruments}, row~8); \S\ref{sec:audit} measures whether a credited name is asserted in the relation the question asks for.

\subsection{Semantic audit of the matcher}\label{sec:audit}
Under a model judge, 233 of 240 sampled credits hold up and the estimated exposed-item rate falls by 2.5 points. The matcher tests for the presence of a gold target's name, not that the answer asserts the right relation, that subject and object are the right way round, or that the answer is not denying the relationship. This section measures, on the released sweep itself, how far that gap moves the numbers.

\paragraph{Frame and sample.} The unit is a (question, model, gold item) triple in the scaffold arm, and the frame is every such unit in the ten-model sweep: 510 questions, ten models, $11{,}360$ units. The sample is drawn only after a gate: regenerating the exposure classification from the frozen question set and the scaffold index must reproduce the released decomposition exactly, and it does ($n_{11}=9{,}865$, $n_{10}=735$, $n_{01}=3$, $n_{00}=757$). Sampling is stratified by matcher category, then within each category over (template $\times$ gold type) cells in proportion to cell size, then over models within each cell, with simple random sampling without replacement at the last stage, under seed 42. The result is \textbf{423 units}, sampled item-model observations rather than answers (one answer can contribute several): 240 of the 9{,}865 credited-and-exposed, 120 of the 735 exposed-and-omitted, 60 of the 757 unexposed-and-uncredited, and a census of all 3 unexposed-and-credited. Exposure is model-independent, so the unexposed strata are 76 distinct items repeated across ten models and are not independent observations.

\paragraph{Adjudicator.} Each unit is put to \texttt{openai/gpt-4.1} at temperature~0 through OpenRouter, with the question, the gold edge, the full gold set, whether the question requires all or any one of it, and the answer. The rubric asks for one of six labels: \emph{correct\_relation}, \emph{name\_only}, \emph{wrong\_relation}, \emph{reversed}, \emph{negated}, \emph{absent}, with a fixed precedence order, plus whether the item is required or an accepted alternative, a confidence, and a verbatim quote. All 423 units returned a parseable verdict, two after a single formatting re-ask. Every call is cached by a \texttt{sha256} of judge model, rubric and prompt, so re-running returns identical verdicts; the whole adjudication cost \$0.93. \textbf{The adjudicator is a model, not a human.} It is cross-family to nine of the ten swept models and the same family as the tenth, \texttt{gpt-4.1-mini}, whose row therefore carries a self-preference risk the others do not; the per-model breakdown is released so a reader can check it, and the blind 120-unit sample and annotation guide for a human pass are released unannotated.

\paragraph{Quote gate.} Every non-\emph{absent} verdict must cite a verbatim span of the answer, and the audit checks that span under the matcher's own normalisation. One policy covers every failed citation and never consults the matcher's own category, so credited and uncredited strata are treated alike. A verdict whose cited span cannot be found in the answer is put back to the judge once, with the rubric and the item unchanged, asking for a span that occurs verbatim; a verdict that still fails to supply one is marked \emph{unresolved} rather than silently kept or silently demoted. Of 281 verdicts required to cite, 252 verify at the first pass, and the 29 that do not are adjudicated under that one policy: 11 are repaired by a re-asked span that verifies, and on the other 18 the judge conceded that no supporting span exists. Repairs and concessions fall along the strata: all 11 repairs are in the credited stratum, 6 verifying character for character and 5 under the matcher's own normalisation; all 18 concessions are in the uncredited strata, 14 in $n_{10}$ and 4 in $n_{00}$. A re-asked span is accepted when it matches the answer character for character, or under the matcher's own whitespace and punctuation normalisation, which keeps the gate no stricter than the policy it replaces; both flags are recorded per unit in the released log. Under a strictly verbatim rule the five normalisation-only repairs would be unresolved instead, all five in the credited stratum and all carrying a \emph{correct\_relation} verdict, and credit precision would read $0.950$ ($228/240$) against the matcher, $0.971$ ($233/240$) for it and $0.970$ ($228/235$) with them excluded. Every rate below is read three ways, which bounds what the unresolved verdicts can do to it: taking each unresolved verdict at its word, counting it as \emph{absent}, and dropping it from its stratum's denominator. Under the acceptance rule as run, credited-item rates are invariant across all three, precision of credit holding at $0.971$; the omission rates move, and are quoted with that spread where they appear.

\paragraph{Precision of credit.} Table~\ref{tab:audit} gives the confusion (upper panel). Of the 240 credited items, $233$ are judged to assert the gold edge: precision of credit $\mathbf{0.971}$ $[0.941,0.986]$. Four ($0.017$ $[0.006,0.042]$) name the target without asserting the asked-for relation, one ($0.004$) places it in the wrong relation, and two ($0.008$) the judge cannot find at all. Roughly nineteen credited items in twenty therefore assert what the question asked, and the residual splits about four to one between bare naming and outright relational error. The unexposed side raises two different questions, which the audit answers with different evidence. The first is whether the matcher hides correct answers it never credited: on 60 unexposed-and-uncredited items the judge finds no hidden correct relational answer ($0$ of $60$, $0.000$ $[0.000,0.060]$), which rules out an undetected-recovery rate above about $6\%$ on this sample. Those units say nothing about false credit, because the matcher issued no credit on them; they are uncredited by construction. False credit is answered only by the second question, and only by the $n_{01}$ census: all three unexposed-but-credited items in the entire frame fail relational adjudication, one naming the target without asserting the edge and two the judge cannot locate at all. Frequency and precision are separate facts here, and both are needed: credit beyond exposure is rare, three units in $11{,}360$, and on this census, which covers every such unit, its precision is zero. The three-item beyond-exposure recovery of \S\ref{sec:tenmodel} therefore survives the audit as noise rather than signal, though three units support no rate estimate.

\paragraph{Correcting the exposed-item rate.} Reweighting each matcher stratum by the judged probability that the answer asserts the gold edge moves the exposed-item rate from the matcher's context-utilisation figure of $0.931$ to an estimated $\mathbf{0.905}$ $[0.884,0.923]$, a difference of $2.5$ percentage points. The two figures are different quantities, not two judges: $0.931$ is the lexical, item-pooled rate at which exposed gold names reappear in answers, and $0.905$ is a frame-weighted estimate of how often the answer asserts the gold edge under the model judge. Neither is whole-answer accuracy. It is an estimated correction under this adjudicator and protocol, not a known overstatement, and three qualifications belong beside it. The adjudicator is a model and the demotion rule is a quote policy, so both the size and the sign of the correction inherit the judge's own errors and that policy's choices. Both strata are sampled rather than censused, and their observations repeat questions and gold targets across ten models, so they are not independent draws. And the denominator is lexically exposed items while the judged quantity, whether the answer asserts the gold edge, is a property of answers, not a measure of whether the shown context was sufficient to produce them. The estimator is the frame-weighted combination of the two judged stratum rates, $R=(9{,}865\,p_{11}+735\,p_{10})/10{,}600$ with $p_{11}=233/240$ and $p_{10}=3/120$, and its interval comes from a stratified cluster bootstrap that resamples \emph{questions} within each stratum, 184 of them in the credited stratum and 59 in the omitted one, recomputing both rates on every draw and recombining them through the same weights, at 10{,}000 draws and seed 42. Resampling questions rather than units carries the dependence between gold items of one question; clustering on model instead of question gives $[0.886,0.924]$, so neither crossed factor dominates. Propagating the two Wilson intervals naively gives $[0.876,0.922]$, which is reported for comparison only and is not a valid interval for this design: it adds bounds instead of resampling the joint distribution, and it ignores the clustering altogether. Across the three readings of an unresolved verdict $R$ moves only from $0.9052$ to $0.9054$, so the correction does not turn on how those units are counted. The gap is small partly because the two error directions offset: credited items are occasionally not asserted, and omitted items are occasionally asserted after all. That offset is a fact about this corpus and this matcher, not a general property.

\paragraph{Omissions.} The unresolved verdicts sit in this stratum, so its rates are the ones that move under the three readings. Of the 120 exposed-but-omitted items, true omissions, the target absent and nothing substituted for it, run from $0.708$ to $0.825$ across those readings ($0.802$ when unresolved units are excluded), and the share where the judge finds the item named at all runs the other way, from $0.225$ down to $0.108$ ($0.123$ excluded). Three ($0.025$ in every reading) are outright matcher false negatives where the judge finds the edge asserted; five ($0.042$) are answers that explicitly \emph{deny} the edge, which the matcher and a true omission cannot distinguish; and between five and two ($0.042$ to $0.017$; $0.019$ excluded) supply an accepted alternative instead. That last figure recounts the compliant-omission correction of \S\ref{sec:tenmodel} from the other side. The deterministic any-collapse rule flags $52$ of $735$ omissions ($0.071$) as compliant; the judge, requiring the substitute to be genuinely asserted rather than merely matched, is stricter. The $52/735$ figure should therefore be read as the upper end of that correction, not its point estimate.

\paragraph{Why the matcher errs.} Three mechanisms are visible, and the first two can be counted over the whole frame without any judge (Table~\ref{tab:audit}, lower panel). \emph{Subject-substring credit}: on 208 credited units the gold title is a substring of the question's own subject name, so merely restating the question earns credit. This is the weakest credit path in the audit: of the seven such units sampled, four are judged \emph{correct\_relation} against $0.971$ overall. \emph{Run-together gold titles}: seven gold titles in the frozen set write their words without spaces (\texttt{DeFi}, \texttt{EdDSA}, \texttt{GraphQL}, \texttt{MultiAccessEdgeComputing}, \texttt{OpenXR}, \texttt{SustainabilityReporting}, \texttt{WebRTC}). They are not unmatchable. Of the 70 frame units carrying one, the matcher credits 58 and omits 12 (Table~\ref{tab:audit}, lower panel), and all 58 are credited by a verbatim substring hit: the model reproduced the run-together form the scaffold put in front of it, so the normalisation's refusal to split the title never bit. The blind spot is the spaced expansion, not the title, and it bites on 10 of the 12 omissions, where the answer does contain the spaced form and the matcher cannot see it (``Sustainability Reporting'' for \texttt{SustainabilityReporting}); the other two do not mention the item under any spelling, and there the matcher is right. The defect therefore costs 10 units of $11{,}360$, from two titles: \texttt{SustainabilityReporting} on one question for nine of the ten models, and \texttt{MultiAccessEdgeComputing} once. The remaining five are conventional unspaced names in their fields, which is why models reproduce them exactly and the matcher never misses them. All seven sampled units carrying the mechanism are judged \emph{correct\_relation}. \emph{Taxonomic attachment}: credited-item precision is lowest on T-TAX, $0.938$ $[0.832,0.979]$ against T-REL's $0.978$ $[0.944,0.991]$, and the wrong-relation case is a model naming a correct ancestor while asserting the wrong attachment point in the hierarchy. A matcher that tests for a name cannot see the difference between a parent and a grandparent.

\paragraph{What this licenses.} The proxy is useful for measuring how much exposed gold survives delivery: under this adjudicator it overstates the relational rate by an estimated $2.5$ points, and its credit on exposed items is precise at $0.971$; on unexposed items it issues credit three times in $11{,}360$, and every one of those three fails adjudication. Its failures are not random: they fall on subject-name substrings, on the spaced expansion of run-together gold titles, and on taxonomic attachment. The proxy does not license reading recall as relational correctness anywhere in this paper, and it does not replace a human pass, which is prepared and not performed.

\begin{table}[tb]\centering\footnotesize
\caption{\emph{Top}: semantic audit confusion over 423 sampled item-model observations (not answers),
matcher category against the verdict of a model judge (\texttt{gpt-4.1}; no human scored these units), as
judged (before the quote gate); \emph{Name} is \emph{name\_only}, the target named without the
asked-for relation, \emph{Wrong} pools \emph{wrong\_relation}, \emph{reversed} and
\emph{negated}, and $n_{01}$ is a census of all three units while the other rows are stratified
samples. \emph{Bottom}: deterministic mechanism census over all $11{,}360$ frame units, needing no
judge; each row counts units carrying a structural property of the matcher or the question set, and
the sampled verdicts then say how often each one changes the answer. Run-together gold titles are
matched when an answer reproduces them verbatim and missed when it expands them into spaced words,
which is why they appear in both the credited and the omitted columns; the blind spot is a property
of the frozen question set rather than of any model.}
\label{tab:audit}
\setlength{\tabcolsep}{3.2pt}
\begin{tabular}{lrrrrr}
\toprule
Matcher category & Correct & Name & Wrong & Absent & $n$ \\
\midrule
$n_{11}$ exp., credited   & 233 &  4 & 1 &  2 & 240 \\
$n_{10}$ exp., omitted    &   3 & 24 & 7 & 86 & 120 \\
$n_{01}$ unexp., credited &   0 &  1 & 0 &  2 &   3 \\
$n_{00}$ unexp., omitted  &   0 &  6 & 2 & 52 &  60 \\
\bottomrule
\end{tabular}

\vspace{4pt}
\setlength{\tabcolsep}{3.2pt}
\begin{tabular}{lrrrr}
\toprule
Mechanism & $n_{11}$ & $n_{10}$ & $n_{01}$ & $n_{00}$ \\
\midrule
Gold title inside the subject name & 208 &   2 & 0 &  0 \\
Credit via the $\geq0.8$ overlap path &  28 &   0 & 3 &  0 \\
Gold title runs words together     &  58 &  12 & 0 &  0 \\
Gold title is a single word        & 1862 & 128 & 0 & 80 \\
\midrule
\emph{Denominator}                 & 9865 & 735 & 3 & 757 \\
\bottomrule
\end{tabular}
\end{table}

\subsection{Ten models: large recall uplift, every model below the copy baseline}\label{sec:tenmodel}
Every model gains substantially from grounding and every model stays below the copy baseline. Table~\ref{tab:sweep} applies the paired design to ten models from eight developer families under one fixed configuration. Raw$\to$scaffold uplift ranges from $+0.57$ to $+0.78$, while scaffold recall spans 0.897--0.942 against the model-independent ceiling of \textbf{0.964}, so raw uplift alone does not distinguish delivery from exposure.

\begin{figure*}[t]\centering
\begin{tikzpicture}
\begin{axis}[xbar,width=13cm,height=6.6cm,bar width=9pt,xmin=-0.098,xmax=0.052,
  xlabel={Gain over copy (scaffold recall $-$ copy ceiling)}, enlarge y limits=0.09,
  symbolic y coords={Gemini 2.5 FL,GPT-4.1-mini,Mistral 24B,Qwen 72B,Llama 70B,DeepSeek,GLM-4.6,Haiku 4.5,Gemini 3.5 FL,Gemini 3.7 F},
  ytick=data, xtick={-0.08,-0.06,-0.04,-0.02,0}, clip=false,
  every axis plot/.append style={fill=burnt!80,draw=burnt}]
\addplot[fill=burnt!80,draw=burnt,error bars/.cd, x dir=both, x explicit,
  error bar style={burnt!55,line width=0.6pt}, error mark options={rotate=90,mark size=2.2pt,burnt!55}]
coordinates {
  (-0.0674,Gemini 2.5 FL) += (0.0175,0) -= (0.0184,0)
  (-0.0596,GPT-4.1-mini)  += (0.0157,0) -= (0.0170,0)
  (-0.0583,Mistral 24B)   += (0.0162,0) -= (0.0173,0)
  (-0.0503,Qwen 72B)      += (0.0140,0) -= (0.0155,0)
  (-0.0489,Llama 70B)     += (0.0149,0) -= (0.0159,0)
  (-0.0400,DeepSeek)      += (0.0115,0) -= (0.0126,0)
  (-0.0363,GLM-4.6)       += (0.0113,0) -= (0.0127,0)
  (-0.0307,Haiku 4.5)     += (0.0103,0) -= (0.0115,0)
  (-0.0264,Gemini 3.5 FL) += (0.0080,0) -= (0.0093,0)
  (-0.0222,Gemini 3.7 F)  += (0.0065,0) -= (0.0070,0)
};
\node[anchor=west,font=\scriptsize] at (axis cs:0.004,Gemini 2.5 FL) {$-0.067\ [-0.086,-0.050]$};
\node[anchor=west,font=\scriptsize] at (axis cs:0.004,GPT-4.1-mini)  {$-0.060\ [-0.077,-0.044]$};
\node[anchor=west,font=\scriptsize] at (axis cs:0.004,Mistral 24B)   {$-0.058\ [-0.076,-0.042]$};
\node[anchor=west,font=\scriptsize] at (axis cs:0.004,Qwen 72B)      {$-0.050\ [-0.066,-0.036]$};
\node[anchor=west,font=\scriptsize] at (axis cs:0.004,Llama 70B)     {$-0.049\ [-0.065,-0.034]$};
\node[anchor=west,font=\scriptsize] at (axis cs:0.004,DeepSeek)      {$-0.040\ [-0.053,-0.029]$};
\node[anchor=west,font=\scriptsize] at (axis cs:0.004,GLM-4.6)       {$-0.036\ [-0.049,-0.025]$};
\node[anchor=west,font=\scriptsize] at (axis cs:0.004,Haiku 4.5)     {$-0.031\ [-0.042,-0.020]$};
\node[anchor=west,font=\scriptsize] at (axis cs:0.004,Gemini 3.5 FL) {$-0.026\ [-0.036,-0.018]$};
\node[anchor=west,font=\scriptsize] at (axis cs:0.004,Gemini 3.7 F)  {$-0.022\ [-0.029,-0.016]$};
\end{axis}\end{tikzpicture}
\caption{Signed gain over copy for all ten models, negative throughout: a verbatim copy of the shown
context scores higher on this limited recall measure than any model, which does not show that copying gives
better answers (the copy is not scored for concision, usability or precision) and does not test whether
reasoning occurred. Per-model seeded
10{,}000-resample bootstrap 95\% intervals (whiskers); the value column at right gives each point and
interval, rounded independently from the full-precision artefact (points, whiskers and labels are all
generated from it). No model's interval crosses zero, and none exceeds the copy ceiling on the lexical
metric; the spread ($-0.067$ to $-0.022$) puts the models on an exposure-normalised scale that raw
uplift ($+0.57$ to $+0.78$) does not provide. Models nearest zero (top) omit the fewest exposed gold names.
A value near zero is a net and not a statement of fidelity: it is equally consistent with no omissions
and no recoveries, and with omissions and recoveries that cancel. Only the counts of
Table~\ref{tab:decomp} separate those, and the scale says nothing about the precision or relational
correctness of what each model returns.}
\label{fig:gain}
\end{figure*}
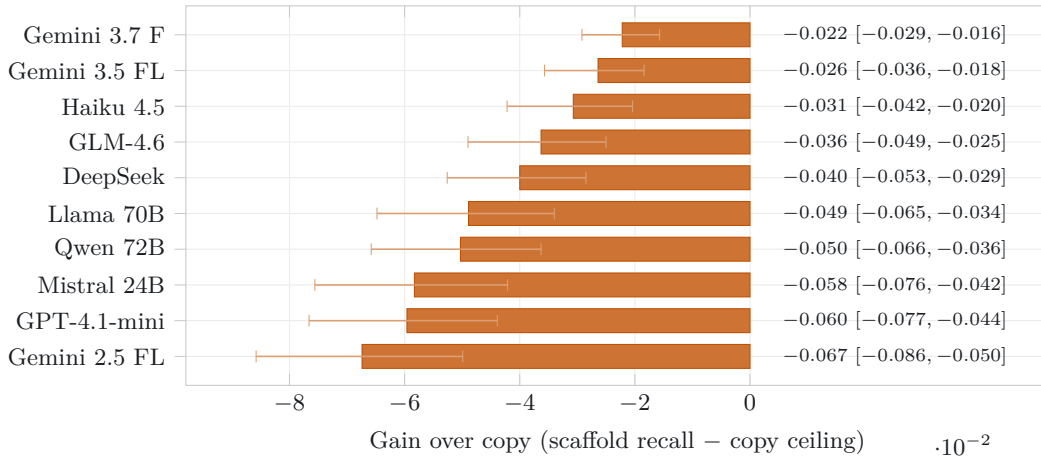

Signed \textbf{gain over copy} is negative throughout (Figure~\ref{fig:gain}), from $-0.067$ for Gemini~2.5~Flash-Lite to $-0.022$ for Gemini~3.7~Flash. Because $g_m=R_m-0.964$, ranking by gain equals ranking by scaffold recall: gain adds an interpretable scale, not a new ordering. Recent models sit nearest the copy baseline, while Gemini~2.5~Flash-Lite, GPT-4.1-mini and Mistral-Small-24B fall further below, an ordering specific to this task. Two models had transport-level truncations under the fixed 2048-token budget. GLM-4.6 had 174 truncated and 5 retried, with the rest scored as received and its row treated as lower-confidence; Gemini~2.5~Flash-Lite had 16 truncated. The 174 GLM-4.6 rows substantially qualify any reading of this table as a model ranking: the configuration is a valid fixed operating point, not a clean comparison of capability.

\paragraph{What adds information: the exposure/recovery decomposition.} A negative $g$ alone cannot distinguish omission of exposed gold from failure to recover unexposed gold, so each item is classified by scaffold exposure and answer recovery. Across ten models and 11{,}360 gold-item observations (the same 1{,}136 target instances scored under each of ten models, not 11{,}360 independent facts), $n_{11}=9{,}865$ were exposed and recovered, $n_{10}=735$ exposed and omitted, $n_{01}=3$ unexposed and recovered, and $n_{00}=757$ unexposed and omitted (Table~\ref{tab:decomp}). Only \textbf{3 of 11{,}360} items ($0.026\%$) were recovered without exposure, and no model exceeds $n_{01}=1$. With $n_{01}\approx0$, $g\approx-n_{10}/|G|$: the deficit chiefly measures exposed gold names absent from answers, including lexical under-counting of paraphrase, rather than recovery beyond exposure. Context utilisation, $n_{11}/(n_{11}+n_{10})$, ranges from $0.900$ to $0.955$. Roughly one exposed gold name in fourteen is absent from the answer ($735/10{,}600=0.069$, one in $14.4$). That raw count includes omissions that correctly comply with any-one-of questions: 166 of the 735 belong to \texttt{any}-type questions, and 52 of those are compliant, the model having recovered a different accepted alternative that the paper's own scorer credits in full. Excluding them gives $683/10{,}600=0.064$, one exposed item in $15.5$, and the two models with the weakest utilisation account for 24 of the 52. The adjusted figure is a \emph{target-name omission rate}, not a bound on semantic error: the matcher can also credit relationally wrong answers and miss other error types, and the audit of \S\ref{sec:audit} estimates the semantic rate directly. The headline recall figures are unaffected, since they already apply the any-collapse at question level; only the item-level drop rate moves. Utilisation is micro-averaged over exposed items whereas recall averages per-question ratios with varying $c_i$, so the two rankings differ once, llama-3.3-70b having utilisation $0.934$ against deepseek-chat's $0.931$ but recall $0.916$ against $0.924$.

\begin{table*}[t]\centering\small
\caption{Ten models, one fixed configuration, $n=510$ (seed 42), sorted by gain over copy
(scaffold recall $-$ copy ceiling). Grounding raises lexical recall of gold target names substantially
for every model (Raw to Scaffold). The copy ceiling (0.964), the recall a verbatim copy of the shown
context achieves, is a property of the questions and the injected context and is identical across
models; every model scores below it on this recall measure, which scores target names, not answer
quality, and does not test whether reasoning occurred. The interval shown is the domain-clustered block
bootstrap (the wider, primary one).}
\label{tab:sweep}
\begin{tabular}{lccccc}
\toprule
Model & Raw & Scaffold & Uplift [clustered 95\% CI] & Ceiling & Gain \\
\midrule
Gemini 3.7 Flash      & 0.375 & 0.942 & $+0.567$ $[+0.524,+0.611]$ & 0.964 & \gainneg{-0.022} \\
Gemini 3.5 Flash-Lite & 0.323 & 0.938 & $+0.615$ $[+0.565,+0.667]$ & 0.964 & \gainneg{-0.026} \\
Claude Haiku 4.5      & 0.151 & 0.934 & $+0.783$ $[+0.741,+0.827]$ & 0.964 & \gainneg{-0.031} \\
GLM-4.6               & 0.243 & 0.928 & $+0.685$ $[+0.641,+0.727]$ & 0.964 & \gainneg{-0.036} \\
DeepSeek-Chat         & 0.347 & 0.924 & $+0.578$ $[+0.532,+0.622]$ & 0.964 & \gainneg{-0.040} \\
Llama-3.3-70B         & 0.227 & 0.916 & $+0.688$ $[+0.632,+0.738]$ & 0.964 & \gainneg{-0.049} \\
Qwen-2.5-72B          & 0.309 & 0.914 & $+0.605$ $[+0.562,+0.649]$ & 0.964 & \gainneg{-0.050} \\
Mistral-Small-24B     & 0.285 & 0.906 & $+0.621$ $[+0.558,+0.682]$ & 0.964 & \gainneg{-0.058} \\
GPT-4.1-mini          & 0.162 & 0.905 & $+0.743$ $[+0.697,+0.787]$ & 0.964 & \gainneg{-0.060} \\
Gemini 2.5 Flash-Lite & 0.227 & 0.897 & $+0.670$ $[+0.623,+0.716]$ & 0.964 & \gainneg{-0.067} \\
\bottomrule
\end{tabular}
\end{table*}

\begin{table}[tb]\centering\footnotesize
\caption{Exposure/recovery decomposition over the 510-question sweep: the same $1{,}136$ gold target
instances scored under each of ten models, $11{,}360$ observations pooled rather than $11{,}360$ independent
facts. All four counts are given per model. Sorted by context utilisation (Util., $n_{11}/(n_{11}+n_{10})$); $n_{11}$ is exposed gold the answer recovered, $n_{10}$ exposed gold it omitted, $n_{01}$ gold recovered though unexposed and $n_{00}$ unexposed gold not recovered. Each row sums to the 1{,}136 gold target instances. Across all ten models $n_{01}$ never exceeds 1, and is 3 pooled: on this lexical measure, recovery
beyond exposure is negligible, which is a statement about the score and not about how answers were
produced. Counts are pooled over
gold items (micro-averaged), whereas the headline recall and ceiling average per-question ratios
(macro-averaged), so these counts do not reconstruct those numbers directly.}
\label{tab:decomp}
\setlength{\tabcolsep}{3.6pt}
\begin{tabular}{lccccc}
\toprule
Model & Util. & $n_{11}$ & $n_{10}$ & $n_{01}$ & $n_{00}$ \\
\midrule
Gemini 3.7 Flash      & 0.955 & 1012 & 48  & 0 & 76 \\
Gemini 3.5 Flash-Lite & 0.947 & 1004 & 56  & 0 & 76 \\
Claude Haiku 4.5      & 0.946 & 1003 & 57  & 1 & 75 \\
GLM-4.6               & 0.945 & 1002 & 58  & 0 & 76 \\
Llama-3.3-70B         & 0.934 &  990 & 70  & 0 & 76 \\
DeepSeek-Chat         & 0.931 &  987 & 73  & 0 & 76 \\
Qwen-2.5-72B          & 0.924 &  979 & 81  & 1 & 75 \\
Mistral-Small-24B     & 0.917 &  972 & 88  & 1 & 75 \\
GPT-4.1-mini          & 0.908 &  962 & 98  & 0 & 76 \\
Gemini 2.5 Flash-Lite & 0.900 &  954 & 106 & 0 & 76 \\
\midrule
\textbf{Pooled}       & \textbf{0.931} & \textbf{9865} & \textbf{735} & \textbf{3} & \textbf{757} \\
\bottomrule
\end{tabular}
\end{table}

\subsection{Omitted targets: grounding reduces recovery of corpus targets the scaffold did not show}\label{sec:suppression}
This result needs no copy baseline. It is a bare-versus-grounded comparison on exactly the $n_{01}+n_{00}$ items the scaffold does not expose, and it has the most direct operational consequence of anything in the paper. Unaided, the models can reach those items: pooled over the ten models and the 760 unexposed gold-item slots (76 per model), the raw arm lexically recovers $\mathbf{12.1\%}$ of them (92 of 760). Under injection the \emph{same} items are recovered at $\mathbf{0.4\%}$ (3 of 760); per model, seven of the ten recover none of their 76, and pooled across all ten the total is 3, not zero. On these items, recovery without injected context exceeds recovery with it. One candidate explanation is displacement of parametric knowledge by injected context, a documented tendency~\cite{shi2023distracted,yoran2024robust,longpre2021entity}, though an item absent from the scaffold is omitted rather than contradicted, and this design observes the effect without isolating its causes.

\paragraph{Exactly what the tested items are.} The items are gold targets \emph{from this corpus} which the \emph{particular retrieved scaffold} for their own question happened not to expose, assessed under \emph{this} authority instruction, \emph{this} confidence-gated retrieval policy and \emph{this} lexical matcher. ``Absent from the retrieved scaffold'' is not ``outside the corpus'', still less ``arbitrary out-of-domain knowledge'', so nothing here licenses a claim about how a grounded model handles knowledge the corpus does not contain. Obeying an authoritative supplied source can also be the desired behaviour, so this is a cost to price rather than a defect to fix. Three caveats bound the reading: the shared lexical matcher makes both rates surface proxies for paraphrased recovery; a bare-arm hit does not prove purely parametric provenance, since lexical cues in the question can contribute; and the effect is a property of this injection policy rather than of grounding in the abstract. It motivates sufficiency checks, explicit supplementation and abstention as experiments; none is implemented here.

\paragraph{Are the bare-arm hits real?} The 92 raw-arm recoveries are lexical matches, so the gap the audit of \S\ref{sec:audit} opens on the scaffold arm opens here too: a bare answer can name a gold target without asserting the relation the question asked for. All 92 were put to the same relational adjudicator, under the same rubric and quote policy. The judge calls 71 of them \emph{correct\_relation} ($0.772$ $[0.676,0.846]$), 15 a bare naming and 6 absent, with none placing the target in a wrong, reversed or negated relation; three citations needed adjudication and one of those stayed unresolved, which moves the rate no further than $0.780$. Adjudication therefore validates 71 of the 92 bare-arm credits and none of the 3 grounded-arm credits. It costs the bare arm 21 of its 92 and the grounded arm all three, so the direction of the contrast survives on the validated credits; per model, the bare arm's validated count runs from 1 of 1 to 13 of 14. Because lexical negatives were not exhaustively adjudicated in either arm (60 of the 757 grounded negatives were sampled, none of the 668 bare negatives), these counts do not establish full-frame relational recovery rates, and no interval is attached to them. The contrast stated above is between two lexical scores produced by the same matcher on both arms and is unaffected by this audit; what the audit changes is how much asserted parametric knowledge the bare figure stands for.

\paragraph{Scope of the observation.} The figures are computed with the paper's own byte-identical matcher over the released sweep rows (per-model raw recovery 1--14 of 76). Of the two observations this is the more informative: supplying the context is followed by loss of recall the bare model demonstrably has, a stronger statement than grounded recall merely failing to exceed a verbatim copy. The design observes the behaviour without separating its mechanisms, and the wrapper ablation that would separate them is named in the conclusion's future work.

\subsection{Production study: a model-judged quality benefit on a 0--5 scale}\label{sec:live}
Routing requests through the deployed node improves model-judged answer quality. Study~2 asks a different question from the sweep, on a different outcome: does the deployed path improve answers? It holds Qwen3.8-27B and decoding constant on the reference host and varies only whether requests pass through the shipped Loom node (\emph{loom}), with its live retrieval, confidence gating and injection authority, or reach the backend directly (\emph{raw}). Each answer is scored against a reference on the $0$--$5$ rubric of Appendix~\ref{sec:appendix-rubrics} by a model judge, \texttt{openai/gpt-4.1}, cross-family to the candidate (\S\ref{sec:method}); a second family re-judges every pair as a robustness check (\S\ref{sec:judgefamily}). No human scored these answers; a human annotation sample exists only for the audit of \S\ref{sec:audit}, prepared and not performed. This $0$--$5$ scale is not the lexical recall of \S\ref{sec:tenmodel} and not the page-improvement scale of \S\ref{sec:pagejudge}. Table~\ref{tab:live} reports paired results.

Across $n=114$, loom raises judged quality by $\mathbf{+0.27}$ points ($[+0.11,+0.45]$, exact signed-rank $p=0.0023$, matched-pairs rank-biserial $r=+0.589$). The bootstrap interval targets the mean difference while the exact $p$ targets a rank summary of the paired differences, so the significant $p$ is not itself a test of the mean effect; they are complementary readings of the same 114 pairs. There were 24 wins, 8 losses and 82 ties, so the pooled effect rests on the 32 non-tied pairs. The arcane-set estimate is $+0.79$ ($[+0.17,+1.42]$, exact $p=0.032$, $r=+0.670$), which does not survive Holm correction (adjusted $p=0.097$); the thin-set estimate is $+0.30$ ($[-0.03,+0.63]$), positive but non-significant. For general questions, $+0.05$ $[0.00,+0.13]$ lies within the pre-specified $\pm0.25$ margin, a non-regression reading by interval inclusion (\S\ref{sec:ood}); its exact $p$ is $0.5$ on only two non-zero pairs, and with 58 of 60 pairs tied (mean scores 5.0 loom, 4.95 raw) the instrument is near its ceiling and insensitive on this set. The three set-level point estimates ($+0.79$, $+0.30$, $+0.05$) fall in curation-depth order, but their intervals overlap, so the gradient is suggestive rather than established; no interaction test separates a curation-depth effect from other differences between the sets.

Injection engagement is $95.8\%$ on arcane, $100\%$ on thin and $78.3\%$ on general, so the gate's skipping the remaining $21.7\%$ of general questions does not chiefly explain the general-set null. Median injected context is $\approx1{,}200$--$1{,}300$ tokens across sets. Median latency is comparable for arcane, $89.8$\,s (loom) versus $93.4$\,s (raw), and general, $66.0$ versus $65.7$\,s. Thin is slower under loom, $172.4$ versus $145.2$\,s, consistent with full engagement and longer scaffolds.

\begin{table*}[t]\centering\small
\caption{Study~2, the production study: a separate experiment from the recall sweep, on its own outcome.
Qwen3.8-27B, model held constant, loom versus raw serving path, answer quality scored $0$--$5$ against a
reference by a model judge (\texttt{gpt-4.1}, cross-family to the candidate; no human judging). The served
path gives a model-judged quality benefit, reproduced by a second judge family (\S\ref{sec:judgefamily}); the
$+0.79$ arcane estimate does not survive Holm correction. Paired mean difference with seeded
10{,}000-resample bootstrap 95\% CI, two-sided exact conditional signed-rank $p$ (Holm-adjusted across
the three sets), the matched-pairs rank-biserial $r$ (paired effect size, conditioned on non-tied
pairs), and win/loss/tie counts. Complete-case after budget-exhaustion re-runs (\S\ref{sec:results}).}
\label{tab:live}
\begin{tabular}{lcccccc}
\toprule
Set & $n$ & $\Delta$ (loom$-$raw) & 95\% CI & $p$ (Holm) & $r$ & W/L/T \\
\midrule
arcane           & 24  & \gainpos{0.79} & $[+0.17,+1.42]$ & 0.032 (0.097) & $+0.670$ & 10/3/11 \\
thin             & 30  & \gainpos{0.30} & $[-0.03,+0.63]$ & 0.130 (0.261) & $+0.431$ & 12/5/13 \\
general          & 60  & \gainpos{0.05} & $[\phantom{+}0.00,+0.13]$ & 0.500 (0.500) & $+1.000^{\dagger}$ & 2/0/58 \\
\midrule
\textbf{pooled}  & 114 & \gainpos{0.27} & $[+0.11,+0.45]$ & \textbf{0.0023} & $+0.589$ & 24/8/82 \\
\bottomrule
\end{tabular}
\\[2pt]{\footnotesize $^{\dagger}$ The general-set $r$ rests on 2 non-tied pairs (58 ties); read it as
directional only, not as a magnitude.}
\end{table*}

\subsection{Judge-family robustness}\label{sec:judgefamily}
A second judge family reproduces the production result. Study~2's headline is graded by a single \texttt{gpt-4.1} judge, which leaves the concern that it tracks a family-specific grading style rather than answer content. The same 114 paired answers were therefore re-judged from scratch by \texttt{claude-opus-4-6}, a family used nowhere else in the paper, under the identical rubric, blinding and resume protocol. The contrast reproduces and slightly strengthens: pooled $+0.342$ (33 non-tied) against $+0.272$, with the same per-set ordering (arcane $+1.042$ against $+0.792$, thin $+0.400$ against $+0.300$, general $+0.033$ against $+0.050$, still on two to three non-tied pairs). Different provider families limit shared grading style; they do not establish statistical independence.

\subsection{Negative controls: any on-corpus block beats no context; content specificity is not established}\label{sec:controls}
These controls ask which part of the injected block carries the production study's benefit. Each holds model and question fixed and perturbs only the block, fetched verbatim from the LLM-free \texttt{/loom/scaffold} endpoint (\S\ref{sec:ceiling-controls}). The cohort is 57 frozen questions from Study~2, 24 arcane and 33 thin, answered by the local Qwen3.8-27B at temperature~0 and scored $0$--$5$ by \texttt{openai/gpt-4.1} at temperature~0 under the rubric of Appendix~\ref{sec:appendix-rubrics}, verbatim. Two cohorts were run on these questions: the \textbf{five-arm rerun}, which is primary and reported first, and the \textbf{earlier four-arm cohort}, summarised after it.

\paragraph{Five-arm rerun: design.} The rerun generates the \emph{true}, \emph{shuffled}, \emph{masked}, \emph{irrelevant} and \emph{fluent-noise} arms under one retry policy: 4096 completion tokens, with a single retry at 8192 when the answer comes back empty. Every row persists the exact system and user messages, the injected block and its size, the donor pairing for \emph{irrelevant}, the full attempt ledger, and the block's measured exposure. The \emph{loom} and \emph{raw} arms are not regenerated. They are Study~2's live-harness rows, produced under that harness's 1536-token policy with its 4096-token re-runs (\S\ref{sec:results}) and re-judged by the same judge in the same pass, so every contrast is judge-controlled but contrasts against \emph{raw} or \emph{loom} mix retry policies. Holm--Bonferroni correction runs across the whole eleven-contrast family.

\paragraph{Cohort accounting.} Table~\ref{tab:controls} (lower panel) gives the flow from planned to graded. One question, \texttt{arc\_sov09}, has no scaffold, so the five rerun arms attempt 56 of 57 while the live arms attempt all 57. Under the common retry policy, empty output falls to $1.8$--$5.4\%$ per arm, against $35.7$--$50.0\%$ in the earlier cohort, and every non-empty completion is graded: between 53 and 56 per arm. Each contrast is computed on its own pairwise-complete question set.

\paragraph{What the rerun supports.} Every context-bearing arm beats no context after Holm correction: true$-$raw $+0.81$ $[+0.44,+1.21]$, shuffled$-$raw $+0.75$, masked$-$raw $+0.75$, irrelevant$-$raw $+0.63$, fluent-noise$-$raw $+0.60$ and loom$-$raw $+0.57$ $[+0.22,+0.93]$, Holm $p$ from $0.0004$ to $0.0105$. No content-specificity contrast survives: true$-$irrelevant is $+0.15$ $[-0.13,+0.45]$, true$-$fluent-noise $+0.17$ $[-0.08,+0.45]$, true$-$shuffled $+0.02$ and true$-$masked $0.00$, each with Holm $p=1.0$. A common-intersection repeat on the 51 questions graded in all seven arms reproduces the pattern exactly: all six arm$-$raw contrasts survive Holm and no content contrast does. The supported statement is that \emph{any well-formed on-corpus block of this length beats no context on this cohort; whether the true block's specific content adds value beyond such a block is not established at this power}. Non-significance is not equivalence, and the true$-$irrelevant interval is wide enough to include an effect of nearly half a scale point. On the content contrasts the two question sets point in opposite directions: on arcane questions true$-$irrelevant and true$-$fluent-noise are both $+0.61$, with unadjusted intervals excluding zero, while on thin questions both are slightly negative ($-0.20$ and $-0.17$); neither set-level estimate survives Holm within its set, and the split is a hypothesis for a larger study, not a result.

\paragraph{Measured placebo exposure.} The seed-disjoint placebo is not answer-disjoint, and the rerun measures by how much (Table~\ref{tab:controls}, lower panel). Using the paper's matcher against each question's gold proxy (the titles of its true block's seed classes plus its declared topic), the mean exposure of the injected block is $0.86$ for \emph{true} and \emph{shuffled}, $0.29$ for \emph{irrelevant} (maximum $0.75$; zero on 13 of 56 rows) and $0.17$ for \emph{fluent noise} (maximum $0.60$; zero on 24 of 56). Generic class titles recur across an 8{,}146-class corpus, so a block about other entities still carries some gold names by coincidence, and true$-$irrelevant contrasts more exposure with less, not content with no content. \emph{Masked} replaces seed titles and drops to $0.18$, yet true$-$masked is $0.00$ $[-0.23,+0.23]$; the question text names its own subject, which may supply what masking removes, though this cohort does not test that. The judge scores against free-text references rather than this gold proxy, so exposure here is a covariate, not the outcome. \emph{Irrelevant} blocks average 1{,}240 estimated tokens against 1{,}113 for \emph{true}; \emph{fluent noise} is packed to the true block's length (1{,}075).

\paragraph{Wrapper difference.} The live path and the controls both merge the block into a system message but differ in the wrapper text. The live path prepends an authority instruction (``treat it as ground truth for definitions and relationships\ldots where it is not relevant, ignore it''), whereas the control arms receive the bare block. Contrasts among the control arms share one envelope and are not confounded by it. \textbf{Loom$-$true}, $-0.21$ $[-0.47,+0.06]$, mixes the serving path, the missing preamble and the two retry policies, and carries no attribution.

\paragraph{Earlier four-arm cohort.} The first control cohort ran \emph{true}, \emph{shuffled}, \emph{masked} and \emph{irrelevant} at a flat 1536-token budget, persisted only the answers, and lost $35.7$--$50.0\%$ of control-arm completions to budget exhaustion (28 to 36 of 56 completed per arm). Re-judged with the same judge on 2026-09-21, its scores agree with the archived 2026-08-17 scores on 227 of 240 rows exactly and on all 240 within one point. Across its nine-contrast family only loom$-$raw survives Holm ($+0.59$ $[+0.24,+0.94]$, $p=0.012$); true$-$irrelevant is $+0.39$ $[0.00,+0.87]$ ($n=23$), and a common intersection of 20 questions leaves nothing significant. Its injected blocks were not stored, so its placebo exposure cannot be recovered. It agrees with the rerun on direction, and the rerun supersedes it on power and completeness.

\paragraph{Scope.} These controls test a single local model judged on a $0$--$5$ rubric, whereas the ten-model sweep of \S\ref{sec:tenmodel} is a lexically scored study on a different cohort; neither inherits the other's interpretation. Neither cohort compares the scaffold with strong flat-text retrieval, so neither bears on whether graph structure beats plain text.

\begin{table*}[t]\centering\small
\caption{Five-arm negative-control rerun (57 questions, Qwen3.8-27B, $0$--$5$ answer quality scored
by a model judge, \texttt{gpt-4.1}). Any well-formed on-corpus block beats no context; no contrast
testing whether the true block's specific content matters survives correction, which is
non-establishment at this power, not equivalence. No arm is a flat-text retrieval baseline, so the
graph-versus-plain-text question is untested here. \emph{Upper panel}: pooled arcane\,$+$\,thin
contrasts, paired mean difference with a seeded 10{,}000-resample bootstrap 95\% CI and
Holm--Bonferroni correction across the eleven-contrast family; $^{\dagger}$ marks contrasts surviving
Holm; each contrast uses its own pairwise-complete question set. The loom and raw arms are Study~2's
live-harness rows under the 1536-token policy, so contrasts against them mix retry policies.
\emph{Lower panel}: per-arm flow and the measured gold exposure of the injected block (mean over
56 rows; zero-exposure rows of 56); the placebo arms carry non-zero exposure, so true$-$irrelevant is
not a content-versus-no-content contrast.}
\label{tab:controls}
\begin{tabular}{lcccc}
\toprule
Contrast & $n$ & $\Delta$ (0--5 scale) [95\% CI] & Holm $p$ & W/L/T \\
\midrule
true $-$ raw          & 52 & $+0.81$ $[+0.44,+1.21]^{\dagger}$ & 0.0015 & 25/5/22 \\
masked $-$ raw        & 53 & $+0.75$ $[+0.32,+1.21]^{\dagger}$ & 0.0040 & 26/6/21 \\
shuffled $-$ raw      & 52 & $+0.75$ $[+0.37,+1.15]^{\dagger}$ & 0.0027 & 23/6/23 \\
irrelevant $-$ raw    & 52 & $+0.63$ $[+0.38,+0.92]^{\dagger}$ & 0.0004 & 24/4/24 \\
fluent noise $-$ raw  & 53 & $+0.60$ $[+0.23,+0.96]^{\dagger}$ & 0.0015 & 26/5/22 \\
loom $-$ raw          & 54 & $+0.57$ $[+0.22,+0.93]^{\dagger}$ & 0.0105 & 24/7/23 \\
true $-$ fluent noise & 53 & $+0.17$ $[-0.08,+0.45]$ & 1.000 & 10/4/39 \\
true $-$ irrelevant   & 53 & $+0.15$ $[-0.13,+0.45]$ & 1.000 & 12/8/33 \\
true $-$ shuffled     & 52 & $+0.02$ $[-0.21,+0.25]$ & 1.000 & 7/7/38 \\
true $-$ masked       & 53 & $\phantom{+}0.00$ $[-0.23,+0.23]$ & 1.000 & 6/7/40 \\
loom $-$ true         & 53 & $-0.21$ $[-0.47,+0.06]$ & 0.996 & 5/15/33 \\
\bottomrule
\end{tabular}

\vspace{8pt}
\begin{tabular}{lrrrrrrrr}
\toprule
Arm & Planned & Attempted & Completed & Empty \% & Graded & Mean exp. & Zero-exp. & Tokens \\
\midrule
loom         & 57 & 57 & 56 & 1.8 & 56 & -- & -- & -- \\
raw          & 57 & 57 & 55 & 3.5 & 55 & -- & -- & -- \\
true         & 57 & 56 & 53 & 5.4 & 53 & 0.86 & 0  & 1113 \\
shuffled     & 57 & 56 & 53 & 5.4 & 53 & 0.86 & 0  & 1112 \\
masked       & 57 & 56 & 54 & 3.6 & 54 & 0.18 & 26 & 1116 \\
irrelevant   & 57 & 56 & 54 & 3.6 & 54 & 0.29 & 13 & 1240 \\
fluent noise & 57 & 56 & 55 & 1.8 & 55 & 0.17 & 24 & 1075 \\
\bottomrule
\end{tabular}
\end{table*}

\subsection{Out-of-domain: no regression detected, equivalence not established}\label{sec:ood}
No out-of-domain regression was detected, but non-regression is established only for the single-model production set. In the five-model judged arm (\S\ref{sec:method}), a cross-family judge scored 60 out-of-domain questions $0$--$5$ against frontier gold. Every harness$-$bare 95\% interval includes zero: all questions $-0.05$ $[-0.11,0.00]$, off-domain $-0.09$ $[-0.25,+0.02]$, and the pre-specified worst case of erroneous injection, off-domain questions on which the confidence gate mis-fired and injected anyway, $-0.23$ $[-0.63,+0.05]$ on 40 pairs. The adjacent and in-domain-general subsets are $+0.01$ $[-0.02,+0.05]$ and $-0.06$ $[-0.14,0.00]$ on 100 pairs each. The gate correctly skipped 60 of 100 off-domain gradings, making harnessed and bare answers identical; detected degradation was confined to over-firing and concentrated on the weakest model. The interval-inclusion result comes from Study~2: general-set $\Delta=+0.05$ $[0.00,+0.13]$ lies within the pre-specified $\pm0.25$ margin. That single-model result does not generalise to the five-model arm, whose worst-case subgroup interval $[-0.63,+0.05]$ extends beyond the margin, so equivalence is \emph{not} established there, only not-detected regression. Non-detection under a gate that engaged on $78.3\%$ of general questions is evidence about the deployed policy, not proof that the gate causes the non-regression, since there is no matched always-inject ablation.

\section{A Production Case Study: Podcast Assertion Extraction}\label{sec:casestudy}
Grounding pulled a model's extracted vocabulary towards the corpus. A production deployment tested
whether scaffold grounding helps a model \emph{write into} the corpus. The node's podcast-ingest pipeline\footnote{The ingest pipeline and the ledger writer of
\S\ref{sec:lifecycle} ship in the VisionFlow agent environment:
\url{https://github.com/DreamLab-AI/VisionFlow}; the graph and its CI-gated builder are the
knowledgeGraph repository already cited.} extracts tiered, source-attributed assertions from
episode transcripts, each carrying \texttt{ontology\_terms} used to locate a destination page;
assertions whose terms resolve to no page are discarded or queued for page proposal, which makes
\emph{term resolution against the live graph} the landing metric. We report a lexical proxy,
normalised string matches against the 8{,}142-class inventory of the generation served to that run (the sweep snapshot has 8{,}146); it is
not a human judgement of assertion quality.

We ran the unmodified production prompt over ten episodes of a technology news podcast
(November 2025--August 2026) across ten arms: the node's local Qwen3.8-27B \emph{through the
Loom} (scaffold injected at the default 1{,}500-token budget, verbatim short-circuit disabled)
and \emph{bare}; and eight frontier or near-frontier cloud models, all bare, because the Loom
grounds only the model behind its own fa\c{c}ade. All arms used direct chat-completion calls
with identical sampling (temperature 0.2, 12{,}288 max tokens). Claude and GPT used a model
router; the others used their providers' APIs. One Loom call produced three assertions for one
episode and remains in the totals.

\begin{table*}[t]\centering\small
\caption{Assertion extraction across ten serving arms, ten episodes of one podcast, one run per arm: a
write-path study with its own outcome, not pooled with the recall or quality studies. \emph{Ground}
is the fraction of emitted \texttt{ontology\_terms} that resolve to an existing knowledge-graph
page, a lexical measure of alignment with the corpus and not of assertion accuracy; \emph{Artif.}\ counts claims that reproduce speech-recognition artefacts as
entity names (e.g.\ ``Opus 48'', ``GPT55''). $n$ is total assertions; Conf.\ is the model's own
mean stated confidence, which is self-reported and not comparable as a quality measure across
families.}
\label{tab:casestudy}
\renewcommand{\arraystretch}{1.15}
\begin{tabular}{@{}lrrrrr@{}}
\toprule
Arm & $n$ & Conf. & Ground & Artif. & s/episode \\
\midrule
\textbf{Loom + Qwen3.8-27B (local)} & 125 & 0.83 & \textbf{0.55} & 0 & 173 \\
Qwen3.8-27B bare (local)            & 170 & 0.81 & 0.17 & 1 & 346 \\
\midrule
Gemini 3 Flash (preview)            & 107 & 0.80 & 0.42 & 1 & 13 \\
GLM-4.7                             & 181 & 0.80 & 0.39 & 7 & 17 \\
DeepSeek v4 Flash                   & 131 & 0.86 & 0.28 & 3 & 66 \\
Claude Opus 4.8                     & 171 & 0.75 & 0.27 & 3 & 51 \\
Claude Haiku 4.5                    & 295 & 0.82 & 0.22 & 0 & 242 \\
GPT-5.6                             & 133 & 0.82 & 0.21 & 0 & 23 \\
GLM-5.3                             & 169 & 0.82 & 0.21 & 0 & 33 \\
Claude Sonnet 5                     & 175 & 0.76 & 0.20 & 0 & 50 \\
\bottomrule
\end{tabular}
\end{table*}

With weights, prompt and GPU held fixed, the paired scaffold effect is large: term resolution rises
from 0.17 to 0.55 ($3.2\times$) while assertion count falls, and generation time roughly halves
(173s against 346s per episode), consistent with the scaffold displacing open-ended reasoning.
Grounding narrows the model's vocabulary towards the graph's, as integration requires. Grounded, the
local 27B exceeds every bare frontier arm on this alignment measure (best 0.42). This does not establish that frontier models are worse extractors: several produce more
assertions, and assertion quality is not judged here. Without the graph's vocabulary, however, their
concept labels miss it and their assertions do not land. Fidelity to noisy input matters too: several
arms reproduce speech-recognition errors as entity names (7/181 for GLM-4.7, 3 each for DeepSeek and
Opus 4.8), whereas the Loom-grounded Qwen arm emitted none and the bare Qwen arm one, and corrupted
names damage page matching and downstream dedup fingerprints. All of this is
\emph{single-configuration estimation}: one podcast, ten episodes, one run per arm and one lexical
scorer give a point estimate from one draw, with no sampling variance and no transfer test. The
paired Loom-versus-bare contrast is the least exposed to that limit because both arms share the draw.
The table is evidence about where extracted knowledge lands, not a model-quality leaderboard.

\subsection{Judged page integration: every arm degraded the page}\label{sec:pagejudge}
Term resolution does not show whether a page improves, so for each arm we applied its matched
assertions to sandbox copies of target pages using the production integration mechanic.
A cross-family judge (Gemini~3.1~Pro, temperature~0, blind A/B with seeded order randomisation,
un-blinded at analysis) scored each before/after pair. A second judge, GPT-5.6, re-scored a
20\% subset with 93\% verdict agreement. Every arm degrades judged page quality (means $-1.04$
to $-0.44$ on the judge's $-2..2$ improvement scale, which is separate from Study~2's $0$--$5$ answer
scale); the two arms adjacent to a judge's family are flagged in the released analysis. The before/after
judge caught this failure, which neither term resolution nor the copy baseline could have.

Insertion mechanics alone do not explain the result. A section-splice rewriter emits only an
anchored section edit, which deterministic code applies fail-closed, preserving the original by
construction. Prompt optimisation over four variants on a dev split, validated on held-out pages
under two rubrics (the second written post hoc to credit informativeness
explicitly), left the best variant negative under both judges ($-0.40$ dev, $-0.90$ held-out).
Persistence across mechanic, rubric and judge suggests a structural problem: narrow, newly extracted
facts dilute a mature curated page.

Under a pre-declared decision rule, the integration phase was disabled while extraction and
verification continue, and the redesign addresses \emph{placement} rather than phrasing. That is an
engineering decision taken on this evidence, not a demonstration that the redesign works. A blind before/after judged comparison identifies a value-destroying step before it runs at
scale. The copy-ceiling name-completeness gate of \S\ref{sec:analysis} cannot, because it
grades whether a source exposes the answers to the questions it should answer, not whether inserting
that source into a mature page improves the page. The two are complementary halves of one pre-filter (\S\ref{sec:lifecycle}), and this case exercises
only the judged half.

\paragraph{What the judged gate can and cannot rule out.} Contextual judging is itself error-prone~\cite{contextualjudgebench2025}, so the degradation is measured with an imperfect instrument. Yet the most relevant known bias would mask rather than manufacture it: judges favour text labelled as refined~\cite{calmjudgebias2024}, and our pairs are blind with seeded order randomisation. The dilution effect, whereby non-diagnostic detail weakens a judgement built on diagnostic evidence~\cite{nisbett1981dilution}, is consistent with narrow extracted facts eroding a mature curated page, though style, placement, length and truth of the inserted content are not separated here.

\section{Write-Path Lifecycle}\label{sec:lifecycle}
\begin{figure*}[t]\centering
\includegraphics[width=\textwidth]{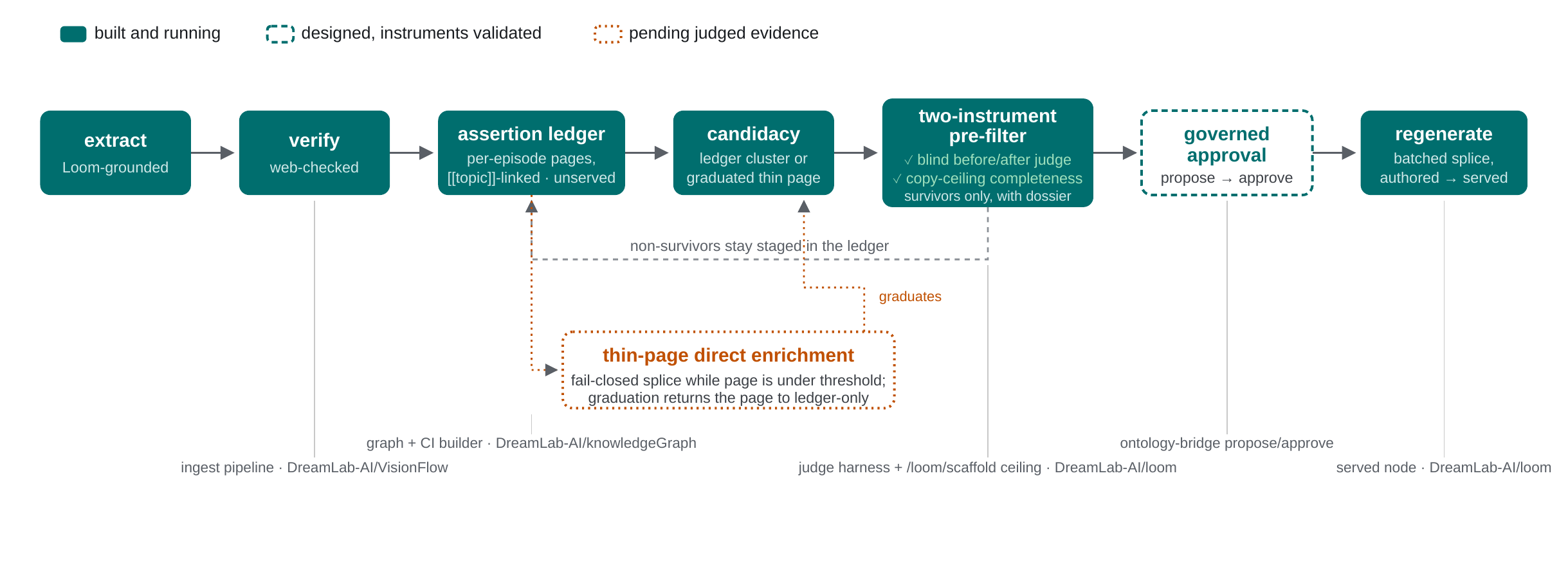}
\caption{Write-path lifecycle and the open-source component carrying each stage, adopted after the
judged integration step degraded pages in every arm (\S\ref{sec:pagejudge}): an operational failure that
recall could not show. The redesign is an engineering decision and is not itself measured.
Landing is separated from serving: assertions stage on unserved
ledger pages, candidates pass the two instruments this paper develops, and only governed,
regenerated content enters the served corpus. The thin-page branch is conditional on its own
judged evidence and self-terminates: an enriched page that crosses the thinness threshold
returns to ledger-only treatment.}
\label{fig:lifecycle}
\end{figure*}

In place of the disabled integration phase (\S\ref{sec:pagejudge}), the write path is built from this paper's two
instruments. Extracted, verified assertions land on
per-episode \emph{ledger pages} inside the graph, written by the same VisionFlow ingest pipeline
that produced the case study above, and wikilinked to the topics they concern: readers see them
immediately through linked references, curated prose is never edited, and, because ledger pages
carry no ontology markup, nothing enters the served corpus by default. Landing is thereby separated
from promotion. A topic whose ledger accumulates evidence, or a thin page whose additions have been
judged to improve it, becomes a \emph{candidate}; candidates pass an automated pre-filter consisting
of the two instruments this paper developed, the blind before/after quality judgement of
\S\ref{sec:pagejudge} and the copy-ceiling name-completeness gate of \S\ref{sec:analysis}.
Survivors reach the graph's existing governed proposal queue\footnote{The governed propose/approve
write path is the ontology-bridge component of the same stack; the serving fa\c{c}ade whose scaffold
and copy-ceiling endpoints the pre-filter reuses is the Loom repository already cited.} as scored
dossiers with provenance down to assertion fingerprints, and approval triggers batched section
regeneration, which is the point at which content acquires ontology markup and becomes servable.
Direct enrichment of thin pages was evaluated as an alternative landing zone and rejected on its
judged evidence: its mean stayed negative across twenty pairs ($-0.65$ and $-0.60$ under two
rubrics), and the section-splice mechanic could not apply at all on thirteen of them for want of
section anchors, leaving the ledger as the single landing zone. Figure~\ref{fig:lifecycle} draws the
path and the open-source component carrying each stage.

As architecture, the lifecycle puts this paper's two instruments to work: the corpus is the
asset, every write to it is measured before it is trusted, and the same ceiling that audits the
serving layer gates admission to it. The ledger-and-served split follows the Event Sourcing / CQRS
materialised-view pattern~\cite{fowler2005eventsourcing} \emph{at the assertion layer}: the ledger is
the append-only source of truth for extracted assertions and the regenerated sections built from it
are disposable projections, which is why a diluting projection can be discarded without data loss.
Mature curated prose predates the ledger, is retained as a source in its own right and is never
rebuilt from it, so the pattern describes the assertion layer and not the whole corpus. Promotion is
idempotent by assertion fingerprint, so the at-least-once delivery inherent to such pipelines cannot
double-promote. Nothing in this section is a measurement: it is the architecture the two measured
gates were used to choose.

\section{Reasoning-Budget Exhaustion}\label{sec:practitioner}
A reasoning model behind an injected scaffold can spend its whole completion allowance thinking and
return nothing. This hazard explains several reporting deviations (\S\ref{sec:results}) and
constrains the token budgets used throughout. At \texttt{max\_tokens}=1536, 42 of 234 loom/raw completions in Study~2 exhausted the
budget \emph{during reasoning} and returned no text; the earlier four-arm control cohort, left at 1536,
lost $36$--$50\%$ of completions the same way (irrelevant $50.0\%$, true $42.9\%$, masked $41.1\%$,
shuffled $35.7\%$; the ordering does not track scaffold disorder, and no per-length analysis was
run, so we report the rates without a monotonic claim). Rerun at 4096 tokens with one retry at 8192,
the same questions lost $1.8$--$5.4\%$ per arm (\S\ref{sec:controls}). The mechanism is a
prompt\,$\times$\,reasoning-budget interaction, with input and output capacity kept distinct: the
scaffold consumes the \emph{context window}, while the reasoning trace and the answer consume the
configured \emph{completion allowance}; a longer or noisier prompt induces a longer reasoning trace,
and a reasoning model that exhausts the allowance mid-trace emits nothing rather than a truncated
answer. Two consequences follow. First, the completion allowance must cover reasoning plus answer, not
answer alone: extraction in the case study uses $12{,}288$ tokens and the Study~2 re-runs $4096$, and
any sub-$2000$-token allowance is unsafe for a reasoning backend behind a non-trivial scaffold. Second,
empty completions must be detected explicitly rather than scored as wrong, because the first-attempt
empty rate is itself a deployment failure signal ($42$ of $234$, $17.9\%$) and complete-case results
measure answer quality conditional on an answer being produced, not service effectiveness.
Temperature~0 does not guarantee deterministic remote inference: ``deterministic'' here refers to
scoring, scaffold generation and fixed decoding settings, not to provider-side reproducibility.

\section{Discussion: What Changes in an Engineer's Decisions}\label{sec:analysis}
\paragraph{What the recall evidence attributes.} In the sweep, names the context exposed account for grounded recall: it is achievable from exposure alone, so this evaluation cannot attribute it to anything further. That is a statement about what the instrument can credit, not about what the models did. One scope condition holds throughout: the $\approx0.92$ grounded recall is for questions phrased in the graph's title vocabulary, and under paraphrase the copy baseline itself falls to $0.328$ (\S\ref{sec:paraphrase}), so the headline is the high end of a distribution rather than a single operating point.

\paragraph{Report the accounting beside quality, not instead of it.} Raw uplift is large and consistent across the sweep, while gain over copy is small and uniformly negative: every model improves on its bare baseline, none exceeds a verbatim copy of its context on recall of target names, and they differ chiefly in how many exposed names their answers omit. The accounting adds an exposure-normalised reference point, not a ranking of answer quality. On a different outcome, the production study found a model-judged quality benefit that the recall figures neither predict nor contradict. A generated answer can justify its cost at lower lexical recall than a copy if it improves task success, clarity or time-to-answer over handing the user the block; those outcomes need direct measurement, and beyond Study~2's model-judged quality this paper measures none of them.

\paragraph{What the accounting does not price.} An organisation builds a node of this kind not
because a model reasons better with a graph attached, but so that its own maintained knowledge is
answerable through a model it can replace, with control over which source is consulted, which version of that source is current, and
by what route a new assertion becomes authoritative. Selecting that material, keeping it current,
governing what may be added to it and delivering the right part of it to a given request are
substantial system responsibilities, and most of the engineering in this stack is spent on them
(\S\ref{sec:ecosystem}, \S\ref{sec:lifecycle}). Exposure accounting identifies which part of an
observed result those responsibilities already explain. It does not measure what they are worth,
and a negative gain over copy is no argument against them: a system can be useful
because it reliably makes maintained knowledge available through a model, whether or not its recall
exceeds what a copy of the shown material would score. Grounding gave a large recall improvement and,
separately, a model-judged quality improvement. A graph may contribute through retrieval, organisation,
versioning and governance as well as through inference, and this study does not isolate the value or
the cost of any one of them. This evidence supports three statements with confidence: versioned curated material is served through a replaceable model interface
(\S\ref{sec:system}); the scaffold supplies the gold target names for the large majority of
in-domain questions (\S\ref{sec:indomain}); and each answer carries source metadata naming the
generation it was served from. Three others still need evidence: that this path is cheaper than an
alternative at matched quality; that ontology structure outperforms simpler retrieval over the same
corpus at an equal budget; and that the attribution metadata is accurate at
the level of an individual answer (Table~\ref{tab:instruments}, rows~7 and~8).

\paragraph{Reading the controls.} Both control cohorts show the served block, and any well-formed on-corpus block of similar length, beating no context; in neither does a content-specificity contrast survive correction (\S\ref{sec:controls}). These graded results therefore do not establish that the scaffold's specific content carries the effect beyond what such a block supplies. This is non-establishment under limited power, with placebo arms that carry measured non-zero exposure, not evidence that content is interchangeable; it matches the shape of the noise confound that~\textcite{cuconasu2024poweofnoise} warn evaluations to control for. Seed-IRI disjointness still matters as method, because an unconstructed placebo carries donor vocabulary and flatters the baseline, but it is a construction guarantee rather than answer disjointness.

\paragraph{An open question: graph against text.} Table~\ref{tab:instruments} leaves two rows unmeasured, and the first most constrains any architectural reading, because a bare model is not the baseline a curated scaffold competes against. The comparison that would bear on investment in graph-structured serving is a matched one: the same questions and models grounded in strong flat-text retrieval over the same corpus, and in direct graph query over the closure, at an equal token budget, with answers scored for task success. A preregistration specification for such a study, on a non-public and provenance-documented corpus, is recorded as proposed future work\footnote{PRD-028, under \nolinkurl{docs/design/} in the served-node repository; a proposal, with no experiment run.}; nothing in this paper anticipates its result, and the accounting of curation effort against query volume that an investment decision would also need is not made here.

\paragraph{Consequence 1: compute the copy baseline before choosing an architecture.}
Compute it on representative questions first. A value close to 1 ($0.964$ here) says the answer entity names are exposed in the context; it does not say the required relation is present, or anything about how an answer reaches it. Where it is high, retrieval already supplies most target coverage and a large bare-model uplift is what exposure alone predicts, so the reference belongs beside the quality and system baselines a generative component is judged on. A low value is uninformative on its own: \S\ref{sec:paraphrase} shows this corpus falling from $0.964$ to $0.328$ through phrasing alone, with no change in what composition could recover. Ruling out retrieval, vocabulary, gold and matcher failure is a precondition for reading a low value as evidence that a task needs composition, and only then does graph machinery such as community summaries, multi-hop traversal or GNN soft-prompts have a measured case to answer. Whether published graph-RAG gains were measured in the high-exposure setting can often be checked from their own reported retrieval coverage.

\paragraph{Consequence 2: the instrument doubles as a promotion gate.} Where the model's measured recall is achievable from what the corpus supplies, one might hypothesise that investment should favour upstream curation, whose cost is amortised rather than repaid per query. That stays a hypothesis: this paper measures neither curation effort against query volume nor a quality-matched alternative serving path. The instrument supports something narrower. Automated extraction candidates, such as blocks proposed by OntoCast~\cite{ontocast} (treated here solely as a third-party upstream producer), can be evaluated through the copy ceiling of the questions they answer. A block that drives that ceiling towards 1 is \emph{name-complete under the matcher}: the answer strings for those questions are present in it. That reading is weaker than completeness of the answers themselves, because the audit of \S\ref{sec:audit} shows name presence and asserted relation coming apart on roughly one credited item in sixty. This gate tests name-completeness of the source, not positive gain over it, since requiring an extraction to \emph{exceed} what its own source exposes would penalise one that reproduces its source exactly. It is necessary but not sufficient: term-stuffing answer names inflates the ceiling while leaving a block incoherent or wrong, so exposure coverage must complement the existing CI, consistency and merge-review gates.

\paragraph{Consequence 3: decisions the operational studies changed.} Four design changes follow from results recall could not show. The semantic fallback's trigger must be keyed to retrieval confidence or retriever disagreement rather than to absence (\S\ref{sec:paraphrase}). An authoritative scaffold should be expected to reduce recovery of corpus targets it omits, which makes sufficiency, supplementation and abstention the experiments to run next (\S\ref{sec:suppression}). A write path should be judged on the page it changes: the page-integration step degraded every page in the judged sample and was disabled (\S\ref{sec:pagejudge}). And a reasoning backend needs a completion allowance covering reasoning plus answer, with empty completions counted as failures (\S\ref{sec:practitioner}).

\section{Companion Work}\label{sec:companion}
A companion note\footnote{\texttt{docs/research/companion-routing/} in the served-node repository, with its corpus, per-run reports, embedding caches, analysis scripts and manifest under \texttt{uplift-results/routing/}. That note states its own release gaps: per-item rows survive for one of its three engines.} measures a different instrument on a different task: a typed-decision seam in an agent runtime, where a router must choose one of 115 exposed skill rubrics or decline. A choice over shown options has no natural no-op extractor, so its baseline is constructed from a judge-free ranker with an oracle-tuned decline rule. That constructed quantity is ranker-relative: it measures a judge's advantage over a particular baseline procedure, not the fraction of gold a fixed text exposes. What carries over is the reporting discipline: a choice task whose gold sits among its shown options should carry a fair baseline beside its accuracy, tuned to a real optimum and read by population rather than in aggregate. Four lessons from that note bear on any such evaluation: a coarse threshold grid stepped over the ranker's own optimum and so overstated the
judge's advantage against it; exclusion clauses indexed as positive evidence made turns score
highest on the very option written to reject them; aggregate accuracy concealed that the judge's
entire advantage sat on boundary and late-clause turns and was nil on ordinary picking; and
systematic ranker--judge disagreement located a labelling error in a corpus its authors had already
reviewed.

\section{Limitations and Threats to Validity}\label{sec:limits}
\begin{itemize}[topsep=2pt,itemsep=2pt,leftmargin=1.4em]
\item \textbf{Recall rewards a verbose copier.} The Study-1 metric is recall against graph-derived gold, so a model emitting the entire scaffold verbatim, gold together with arbitrary fabrications, would score at the copy ceiling. A higher recall score is therefore not a better answer. Precision against fabrication and the accuracy of the attribution metadata are unmeasured throughout.
\item \textbf{The audit's adjudicator is a model.} Every verdict in \S\ref{sec:audit} is \texttt{gpt-4.1}'s, not a human's, and that model is the same family as one of the ten swept models. The quote gate and the deterministic mechanism census bound how far a judge's own errors can move the estimates, but they do not replace human annotation; the blind sample for that pass is released and unannotated.
\item \textbf{Corpus provenance.} The corpus is public and almost entirely LLM-derived under the author's direction; its documented gates check consistency and entailment, not factual accuracy, and no systematic human content review is documented. Public availability does not establish that any model saw it in training, synthetic authorship does not establish novelty or reliability, and this study measures neither. The evaluated snapshot (2026-08-15) is not bound to the sweep rows by a recorded identifier: the link rests on the scaffold index's generation timestamp. Its page count (8{,}138) is the build's reported figure; no independent recount reproduces it exactly.
\item \textbf{Single corpus, and dependent questions.} Both studies use one ontology in one broad domain. The instrument transfers; the magnitudes (a $0.964$ ceiling, a $+0.27$ live lift) do not. Questions generated from one graph cluster by class, domain and shared relation target, so the domain-clustered bootstrap addresses dependence within this corpus, but fifteen domains and one corpus do not establish transfer beyond it.
\item \textbf{Judge coverage.} Study~2's headline uses one GPT-family model judge, cross-family from the candidate, and a re-judge by a third family reproduces the contrast (\S\ref{sec:judgefamily}); ``cross-family'' means a different model provider family from the candidate, not methodological independence from the authors. There is no human judging, no multi-judge panel and no repeatability test across judge seeds.
\item \textbf{The controls are small, mix retry policies against raw, and differ in wrapper on one contrast.} The five-arm rerun grades 53 to 56 answers per arm from 57 questions, which leaves the content contrasts underpowered: a true$-$irrelevant effect of nearly half a scale point is inside its interval, and non-significance is not equivalence. Its loom and raw arms are Study~2's rows under the 1536-token policy with 4096 re-runs, while the scaffold arms ran at 4096 with an 8192 retry, so every contrast against raw mixes policies. The earlier four-arm cohort lost $36$--$50\%$ of control completions, and its surviving rows may describe an easier population than planned. The \emph{irrelevant} arm is about 11\% longer than \emph{true}. The control arms share one bare wrapper, so contrasts among them are not confounded by it; loom$-$true is.
\item \textbf{The placebo is seed-disjoint, not answer-disjoint.} Disjoint seed classes can still share ancestors, relation targets and synonyms. The rerun measures the result: mean gold exposure of $0.29$ for \emph{irrelevant} and $0.17$ for \emph{fluent noise}, so neither is a zero-content arm. That exposure is measured against a gold proxy (seed titles plus topic), not against the free-text references the judge uses. The earlier cohort did not persist its injected blocks, so its placebo exposure is not recoverable.
\item \textbf{Semantic fallback disabled throughout.} HNSW semantic retrieval is off, so every loom arm is lexical only. Enabling it is gated on index recall: the measured value is $0.816$ against $0.87$, a recorded implementation acceptance floor first stated on 2026-08-17 with no recorded derivation or calibration study. That index-acceptance check is distinct from the per-query trigger, which is keyed to lexical absence and is also default off; \S\ref{sec:paraphrase} measures what that trigger would catch.
\item \textbf{Single production model.} Study~2 fixes the model at Qwen3.8-27B, so its magnitude under other backends is unmeasured; the sweep varies models only on the offline scaffold.
\item \textbf{Three excluded questions.} Three thin-set questions still empty in one arm at 4096 are excluded pairwise ($n=114$). The rule is symmetric but the realised missingness was not (loom empty once, raw twice), and a deployment-effectiveness claim would need the empty answers scored as failures.
\item \textbf{Local-model identity.} The local backend is addressed by a compatibility alias (\texttt{qwen3.8-27B}) whose deployed checkpoint is a quantised derivative recorded in the node's deployment notes rather than an attested reference release; provider display names elsewhere are likewise not immutable version identifiers.
\item \textbf{Two exposure figures, not interchangeable.} The macro ceiling, $0.964$, averages per-question exposed fractions; the item-pooled exposed fraction is $(n_{11}+n_{10})/N=10{,}600/11{,}360=0.933$, equivalently $1{,}060$ of $1{,}136$ items per model. Headline recall is macro-averaged and context utilisation is micro-averaged, so the two do not reconstruct each other.
\item \textbf{One seed, one snapshot.} Questions are sampled under seed~42 and served from one corpus generation. The bootstrap intervals estimate uncertainty under resampling of the frozen questions and do not cover corpus, snapshot, prompt or seed variation.
\item \textbf{Reproducibility is pinned, but not complete.} Every table and figure resolves to an artefact listed in the manifest at \nolinkurl{docs/research/gain-over-copy-paper/MANIFEST.md}; this manuscript is tagged \texttt{paper-v10} (\url{https://github.com/DreamLab-AI/loom/tree/paper-v10/docs/research/gain-over-copy-paper}). The five-arm control rerun and the corpus-snapshot note are as of commit \texttt{e7147bec67f1de609ad47bbb06a7aa72d306d9ce}, and the earlier audit, sweep, production and four-arm control artefacts as of commit \texttt{a6d3e41f898c9c18efdf947fc4924ce93c79f4c8}. Two items are absent from that release: the 7.9\,MB scaffold index, which is corpus-derived, so a reader can check every derived number against the released rows but cannot regenerate the scaffolds themselves; and two embedding caches, excluded for size and rebuilt by their own released scripts.
\item \textbf{Equivalence and power.} The out-of-domain reading rests on an instrument near its ceiling: 58 of 60 Study-2 general pairs tie at means of $5.0$ and $4.95$, so sensitivity to deterioration on harder questions is weak and the equivalence claim is confined to that instrument and set. Power figures quoted here are computed against observed effects and are therefore post hoc.
\item \textbf{A constructed baseline is a different instrument.} The copy ceiling applies to a task with a natural no-op extractor. For a task without one, such as a discriminative choice over shown options, a baseline must be constructed from a ranker with a decline rule; that quantity is ranker-relative and is not the exposure scalar of \S\ref{sec:ceiling} (\S\ref{sec:companion}).
\end{itemize}

\section{Conclusion}\label{sec:conclusion}
We built a node that serves a maintained ontology corpus through a replaceable model, and then asked what its evaluation supported. Grounding raised recall of gold target names substantially, from $0.265$ unaided to about $0.92$ across ten models. A verbatim copy of the shown context scores higher still on that limited recall measure ($0.964$), which shows that the names were already in context; it does not show that copying gives better answers, because the copy is not scored for concision, usability or precision. The copy comparison tests what a recall score establishes; it does not test whether reasoning occurred, and it establishes neither its presence nor its absence. A separate paired production study, on its own outcome, found a model-judged quality benefit of $+0.27$ $[+0.11,+0.45]$ on a $0$--$5$ scale, reproduced by a second judge family. The negative controls show any well-formed on-corpus block beating no context and do not establish that the specific content matters, and whether graph-structured grounding beats strong flat-text retrieval is untested, because no matched comparison was run. The corpus is public and largely LLM-generated, which establishes neither that the models saw it in training nor that its content is novel or reliable. Each experiment here uses its own outcome measure, and none is pooled with another.

Five deployment lessons follow. Compute the copy baseline and the four exposure counts before reading a recall uplift, and report them beside quality judgements rather than in place of them. Key a retrieval fallback to confidence or retriever disagreement, not to absence: rephrasing out of the graph's vocabulary collapsed exposure from $0.964$ to $0.328$ while the absence-keyed trigger would have fired on 2 of 506 questions. Expect an authoritative scaffold to reduce recovery of corpus targets it omits ($0.121$ bare to $0.004$ grounded under this wrapper and retrieval policy). Judge a write path on the page it changes: inserting extracted facts degraded judged pages in every arm, and the step was disabled, a failure no recall figure could show. Give a reasoning backend a completion allowance for reasoning plus answer, and count empty completions as failures. None of this is a verdict on knowledge graphs or on model reasoning; it is a record of what these instruments measured on this corpus, and of the decisions they changed.

\paragraph{Future work.} Five experiments would extend the record. \emph{(i)}~A matched comparison of the ontology scaffold against strong flat-text retrieval and direct graph query over the same corpus at an equal budget, scored for task success, on a non-public, provenance-documented corpus (the proposal of \S\ref{sec:analysis}). \emph{(ii)}~A human annotation pass over the 120-unit blind sample already prepared beside the audit, to place agreement and $\kappa$ against the model adjudicator. \emph{(iii)}~A wrapper ablation on the omitted targets of \S\ref{sec:suppression}: hold them fixed, vary the instruction across authoritative, incomplete-context-warning, explicit-supplementation and neutral wrappers, and score by outcome type (omission, contradiction, abstention) rather than right or wrong. \emph{(iv)}~A multi-hop design that varies the route while holding the answer name exposed (Appendix~\ref{sec:appendix-composition}), the instrument the accounting cannot substitute for. \emph{(v)}~A larger control cohort with an answer-disjoint placebo, filtered on measured exposure rather than seed disjointness, and loom and raw arms regenerated under the same retry policy, powered to test the arcane-versus-thin split the five-arm rerun suggests.

\appendix
\onecolumn
\section{Judge rubrics (excerpted)}\label{sec:appendix-rubrics}
This appendix reproduces in full, as run, the production-node semantic answer judge
(\S\ref{sec:method}, behind the $+0.27$ headline), and then summarises the judged-page rubrics. The
answer judge grades a candidate answer against a reference on a $0$--$5$ scale:

{\footnotesize\begin{verbatim}
You are a strict, fair evaluation judge. Grade the CANDIDATE ANSWER against
the REFERENCE ANSWER for the QUESTION, on a 0-5 integer scale:
5 = fully correct and complete vs the reference; 4 = correct, minor omission;
3 = partially correct, no major error; 2 = mostly wrong or one right point;
1 = on-topic but no correct content; 0 = wrong, irrelevant, or fabricated.
Judge CONTENT match to the reference, not style or verbosity. If the candidate
adds correct information beyond the reference, do not penalise it. If it
contradicts the reference, penalise heavily.
Respond with ONLY a JSON object: {"score": <0-5>, "why": "<one sentence>"}.
\end{verbatim}}

\noindent The judged-page gate (\S\ref{sec:pagejudge}) scores a blind before/after page pair on the
same $0$--$5$ shape, returning \texttt{factual\_grounding}, \texttt{relevance}, \texttt{coherence}, a
\texttt{better\_version} verdict and a $-2..2$ \texttt{improvement} field scored for version~B relative
to version~A. Rubric~A judges quality as a reader would, with no indication of how either version was
produced; rubric~B is post-hoc and differs only in framing the reader as
seeking current, accurate knowledge so that informativeness and currency are weighed alongside prose
quality. Rubric~B validates the rubric~A measurement rather than replacing it, and both complete templates,
with their marked data slots, ship with the released harness.

\section{Worked Example: Question, Gold, Paraphrase and Scaffold}\label{sec:appendix-samples}
One item from the sweep makes the copy-ceiling inputs concrete. The question, its gold
target set, and the paraphrase are all verbatim from the frozen sets; the paraphrase is this
question's entry in the vocabulary-mismatch stress set (\S\ref{sec:paraphrase}), accepted on its
first attempt by the gate that requires meaning preservation, no verbatim gold and no seed-title
token.

{\footnotesize\begin{verbatim}
QUESTION (id q0003):
  In the DreamLab knowledge graph, what is the immediate parent concept of
  Affective Computing? Also name up to three broader ancestor concepts.
GOLD (slug/title targets asserted by the graph):
  { human-computer-interaction / "Human Computer Interaction" }
PARAPHRASE (stress-set rewrite; tokens "affective", "computing" banned):
  In the DreamLab knowledge graph, what is the direct parent category of
  the field focused on enabling machines to recognize and respond to
  human emotions? Also, list up to three more general ancestor
  categories.
\end{verbatim}}

\noindent The deterministic scaffold served for this question (budget 1500 tokens, max\_seeds 4,
hops 1, prose off) begins:

{\footnotesize\begin{verbatim}
The following ontology context was retrieved from a curated knowledge graph.
Where it is relevant to the user's request, treat it as ground truth for
definitions and relationships between the concepts it covers. Where it is not
relevant, ignore it and answer normally.

[ONTOLOGY CONTEXT]
## Affective Computing (ai, maturity: emerging)
Affective computing is a branch of artificial intelligence and human-computer
interaction concerned with systems that can recognise, interpret, process, and
simulate human emotions and affective states. [...curated prose continues...]
is-a: Human Computer Interaction; ancestors: AI Research Area,
      Artificial Intelligence
relations: requires: Emotion Recognition, Annotated Training Data, ...
\end{verbatim}}

\noindent Here the gold title ``Human Computer Interaction'' appears verbatim in the scaffold
(as the \texttt{is-a} parent), so the per-item copy ceiling is $c_i=1$: a no-op extractor of the
context would already score the answer. This is the high-exposure case the ceiling reports.
Two properties of this frozen item guard against over-reading the example. The template asks for
broader ancestors as well, but the graph asserts gold here as the immediate parent alone, so the ancestor
names visible in the scaffold are context rather than gold. And although the engine ran with
\texttt{prose~off}, each served block always carries its curated definition text: the flag controls
additional generated prose, not the block's own content.

\section{Two-Edge Composition: a Protocol}\label{sec:appendix-composition}
This appendix specifies a design, not a result: no quantity below has been measured and no
execution is reported. The protocol is recorded because
exposure accounting cannot stand in for it: a question can require following two relations while
its answer name is already a string in the shown context, in which case exposure is $1$, a perfect
answer scores $1$, and gain over copy is exactly zero (\S\ref{sec:ceiling}).

\paragraph{Question construction.} Mine two-edge chains $A\xrightarrow{r_1}B\xrightarrow{r_2}C$ over
the restriction-encoded relations of the reasoned closure. A chain is eligible when no direct
asserted or inferred $A\to C$ edge exists, when $C$ does not appear in $A$'s served block, and when
the chain admits at most a small bounded set of valid targets. The question names only $A$ and the
two relation types, never $B$: it asks for the endpoint of a two-edge relation, and the controls below test whether both premises are needed.

\paragraph{Arms.} Every context-bearing arm serves the same number of blocks, seed-shuffled, so that
the presence or the position of a block does not itself signal relevance. The arms are: \emph{both}
premises ($A$'s and $B$'s blocks plus seed-disjoint distractors); \emph{$b$-only} and \emph{$a$-only}
single-premise controls, each padded with distractors to the same block count; \emph{masked}, in
which both premises are served with their relation labels replaced by uninformative tokens;
\emph{distractor}, in which every block is seed-disjoint from the chain; and \emph{raw}, with no
context. The $a$-only, distractor and raw arms bound parametric leakage and placebo response; the
$b$-only arm bounds what is obtainable from the answer-bearing block alone.

\paragraph{Scorer and the primary contrast.} Score by the same deterministic surface matcher used
throughout, against the bounded target set, and report precision alongside recall so that
enumeration of the candidate space cannot pass as an answer. Because the $b$-only arm already
exposes the target name verbatim, no copy baseline separates it from the both-premises arm; the
informative quantity is the contrast $\Delta=\textrm{both}-b\textrm{-only}$, which nets out answer
exposure. Read each $\Delta$ against its own headroom $1-b_{\textrm{only}}$: a high single-premise
rate caps the detectable difference, so a flat $\Delta$ at a high $b$-only rate is uninformative
rather than negative evidence.

\paragraph{Decisive controls.} Four manipulations supply evidence that a recall rise on adding a relevant
block cannot. \emph{Entity-preserving edge swap}: rewire $r_2$ so that the correct answer
changes while the served names and block lengths are held fixed; a model that follows the changed
edge to the changed answer is evidence of composition. \emph{Relation reversal}: reverse the
direction of $r_2$ at matched names and length, so that the same entities admit a different correct
target. \emph{Consistent renaming}: replace all chain entities with unfamiliar identifiers
consistently throughout the served blocks and the question, which removes parametric recognition
while preserving the route. \emph{Bounded precision}: cap the valid target set and score precision,
so broad enumeration is penalised rather than credited.

\paragraph{Multiplicity and interpretation.} With one $\Delta$ per model, apply Holm correction
across the panel and report Holm-adjusted $p$-values, with confidence intervals labelled as unadjusted unless a simultaneous-coverage procedure is defined; without it, boundary assignments between
positive, null and negative $\Delta$ are expected by chance alone. What the contrasts can identify
is the effect of adding route information, and of altering it, under this particular context
construction. What they cannot identify is composition ability as a model property: adding $A$'s
block changes distractor competition and answer-selection pressure as well as supplying a missing
premise, the masked arm confounds label removal with a change in block surface form, and any grouping
of models by the sign of $\Delta$ is configuration-specific behaviour rather than a capability
ladder.

\twocolumn
\printbibliography

\end{document}